\documentclass[a4paper,fleqn]{cas-sc}

\usepackage[authoryear,longnamesfirst]{natbib}
\usepackage{float} 
\usepackage{algorithm} 
\usepackage{algpseudocode} 
\usepackage{physics} 
\usepackage{scalerel,amssymb} 
\usepackage{booktabs} 
\usepackage{tikz} 
\usetikzlibrary{matrix,backgrounds} 
\pgfdeclarelayer{myback} 
\pgfsetlayers{myback,background,main} 

\tikzset{mycolor/.style = {line width=1bp,color=#1}}%
\tikzset{myfillcolor/.style = {draw,fill=#1}}%

\def\tsc#1{\csdef{#1}{\textsc{\lowercase{#1}}\xspace}}
\tsc{WGM}
\tsc{QE}

\makeatletter
\renewcommand*\env@matrix[1][\arraystretch]{%
  \edef\arraystretch{#1}%
  \hskip -\arraycolsep
  \let\@ifnextchar\new@ifnextchar
  \array{*\c@MaxMatrixCols c}}
\makeatother

\floatname{algorithm}{Algorithm}

\def\msquare{\mathord{\scalerel*{\Box}{gX}}}

\NewDocumentCommand{\highlight}{O{blue!40} m m}{%
\draw[mycolor=#1] (#2.north west)rectangle (#3.south east);
}

\NewDocumentCommand{\fhighlight}{O{blue!40} m m}{%
\draw[myfillcolor=#1] (#2.north west)rectangle (#3.south east);
}



\newdefinition{definition}{Definition}
\newtheorem{theorem}{Theorem}
\newdefinition{rmk}{Remark}
\newproof{pf}{Proof}
\newdefinition{ex}{Example}

\begin{document}
\let\WriteBookmarks\relax
\def\floatpagepagefraction{1}
\def\textpagefraction{.001}

\shorttitle{Vine matrix representations}    

\shortauthors{D. Pfeifer and E. A. Kovács}  

\title [mode = title]{Matrix and graph representations of vine copula structures}

\tnotemark[tnote mark] 


%

\author[1]{Dániel Pfeifer}[type=editor,orcid=0000-0002-2434-6052]



\ead{pfeiferd@math.bme.hu}

\ead[url]{https://www.ttk.bme.hu/}

\credit{Writer, Deviser of Proofs, Implementation}


\affiliation[1]{organization={Budapest University of Technology and Economics},
            addressline={Műegyetem rkp. 3}, 
            city={Budapest},
            postcode={1111}, 
            country={Hungary}}

\author[2]{Edith Alice Kovács}[orcid=0000-0001-7687-4335]


\ead{kovacsea@math.bme.hu}

\ead[url]{}

\credit{Writer, Conceptualization of this study, Methodology, Mathematical background}

\affiliation[2]{organization={Department of Differential Equations, Budapest University of Technology and Economics},
            addressline={Műegyetem rkp. 3}, 
            city={Budapest},
            postcode={1111}, 
            country={Hungary}}



\nonumnote{}

\begin{abstract}
Vine copulas can efficiently model multivariate probability distributions. This paper focuses on a more thorough understanding of their structures, since in the literature, vine copula representations are often ambiguous. The graph representations include the original, cherry and chordal graph sequence structures, which we show equivalence between. Importantly we also show a new result, namely that when a perfect elimination ordering of a vine structure is given, then it can always be uniquely represented with a matrix. O. M. Nápoles has shown a way to represent vines in a matrix, and we algorithmify this previous approach, while also showing a new method for constructing such a matrix, through cherry tree sequences. We also calculate the runtime of these algorithms. Lastly, we prove that these two matrix-building algorithms are equivalent if the same perfect elimination ordering is being used.

\end{abstract}




\maketitle

\section{Introduction and related work}\label{intro}

Vine copulas have gained popularity in the last decades because of their flexibility in modeling multiple types of dependencies at the same time. The extent of their application was recognized in the last couple of years. To get a feel for the multitude of domains where the vine copulas were applied, we will list some of them.

In  \cite{brechmann2013risk}, the goal was to capture complex dependence structures of seven assets of the Euro Stoxx 50 data, and based on this, accurately assessing financial risk. They conclude that vine copula models provide an accurate and efficient forecast of the Value-at-Risk at high levels. In \cite{kielmann2021stock}, the dynamic, nonlinear dependence and risk spillover effects between BRICS stock returns and the different types of oil price shocks were modeled using an appropriate multivariate and dynamic copula model. The risk was measured using the conditional value at risk, conditioning on one or more simultaneous oil and stock market shocks. In \cite{horvath2020copula}, truncated vine copulas were used for the task of anomaly detection in high dimensions. The method was illustrated on a case study, where a large data set consisting of the performance counters of a real mobile telecommunication network was analyzed. In \cite{nikoloulopoulos2017vine} the trivariate vine copulas were used in a meta-analysis of medical data.

The important work of Kurowicka et. al.  \cite{kurowicka2006uncertainty} contains many important results, such as the  valid correlation matrix construction. This is the first book that relates vine copulas, in the special case of multivariate normal distributions to conditional independence, and gives the relationship between conditional rank and partial correlations on a regular vine. Also in this book, the concept of a Markov tree appears, which defines the probability distribution over a tree.

An excellent explanation of pair copulas with applications in financial returns can be found in \cite{aas_dvine}. Dependence Modeling with Copulas \cite{joe} covers several meaningful results related to the field of vine copulas, including tail inference of copulas.

The article \cite{czado2021vine} reviews the basic ideas underlying the vine copula models, and presents estimation and model selection approaches. For a good introduction to vine copulas and a practical guide on using an R package the reader is referred to \cite{czado2019analyzing}. 

An R-vine copula (regular vine copula) gains its flexibility from its three layers: the R-vine structure, the pair copulas used, and the corresponding parameters.

The present paper is concerned with two aspects related to the structure of the R vine: the graphical representations and its matrix representations. In the vine copula literature the widely used graphical representation was introduced in \cite{bedford2}. Other representations like the cherry tree representation (\cite{cherry1}) are not well known.

In this paper we present more possible graphical representations of the vine copula structure, explain them, show the implications between them and prove a new theorem about the condition for their equivalence; then deduce some useful properties from this equivalence. We will dedicate Subsection \ref{subsection:graphstruct} to the rigorous introduction of graph structures, which was motivated by the fact that the papers which deal with the graph structure of vines and chordal graphs, like in \cite{haff} and \cite{zhu}, some mistakes were published. Here we will mention a few of them.

\vspace{2mm}

In \cite{haff}, on Page 190, after defining a decomposable graph as being a chordal graph, the following appears in the next row: "if a graph $G$ describes a decomposable model the joint probability distribution can be written as follows", and a formula is given. Firstly the decomposable model is not defined yet. Next, in the  formula, we have a set of separators and a set of cliques. A "minimal complete separator of two cliques" is defined earlier at the beginning of Section 3 as the intersection of two cliques. However this is not necessarily a separator. See the following probabilistic decomposable model as an example:

$$(1234)-[234]-(2345)-[345]-(3456)$$

The first and the last cliques' intersection is 34, which is not a real separator between the first and last clusters, and hence it is incorrect to take it into the separator set.

Moreover, the formula is also incorrect. The marginal density functions of the separators in the denominator should either be raised to a power $\nu_S$ which is equal to the number of times separator $S$ appears in $G$, or instead of a set of separators, a collection of separators should be used - which accounts for identical separators.

Later on Page 191, the definition of a junction tree is also ambiguous. Firstly, the clusters of a junction tree are the maximal cliques of the input graph, and by their definition, "the edges are the separators that connect the cliques", however these separators are not correctly defined (see the above counterexample).

Secondly, on Page 191, it is false, that a "junction tree is a tree-structured representation of an arbitrary graph". It is a representation for chordal graphs only. For example the graph that has exactly $4$ nodes arranged in a cycle has no junction tree representation. Its maximal cliques are the four cliques of sizes $2$, all individually connected by a $1$-element separator, arranged in a cycle - therefore the outcome cannot be a junction \textit{tree}, it can only be a "junction \textit{graph}".

In \cite{zhu}, the notion of a clique tree is introduced on Page 3. In the definition they used the concept of a "unique path", they did not define. Moreover, the notion of a "path" is not defined in a "graph defined by sets". Also related to this definition, they try to define the running intersection property as a characterization of the so-called clique tree. However on Figure 1, on the lower right side, the image is not a clique tree, because element 1 contained in two clusters of the clique tree is not contained by the intermediary cluster.

Having said this, we think it is important to give a rigorous introduction to the graphical structures used in this topic, which we will do in Section \ref{section:prelim}.

\vspace{2mm}

In order to count the number of all possible vine copula structures
Oswaldo Morales Nápoles (\cite{napoles}) gave a matrix representation of the graphical vine
structure. Their algorithm associates a non unique matrix to the graph
structure. We will analyze the algorithm, and give conditions for the uniqueness.
Using the cherry tree representation we will give a brand new algorithm which
constructs the matrix, tree by tree, i.e. row by row while O. M. Náploes'
algorithms constructs the matrix column by column. We will prove a new theorem which assures
that the two algorithms will construct the same matrix starting from the
same graph structure as an input.

Throughout the paper we will visualize the concepts and the steps of the
algorithm for a better understanding.

The paper contains 5 sections. In the second section we will define and discuss
the mathematical object that will be used. That section contains the graph
theoretical concepts, the copula function in general and the vine copula as
a special case of copulas. The third section contains alternative graphical
representations of vine copula structures and new results that prove relations between them.
In the fourth section, we will introduce a new algorithm for constructing a
matrix, which encodes a vine structure, and discuss the algorithm introduced
by O. M. Nápoles. We give the main result of the paper, which is a theorem that ensures their equivalence.

We will also introduce a new concept, the vine structure's perfect elimination ordering. This will be used heavily throughout our algorithms. We will prove that such an ordering exists, and it is a perfect elimination ordering of all trees in the vine structure.

We hope that through this paper the reader will get a better understanding
of vine copula structures, and as such a better understanding of the matrix
encoding algorithms.

\section{Preliminaries}\label{section:prelim}

This section contains $3$ subsections, which contain the mathematical objects that the following sections rely on.



\subsection{Graph structures}\label{subsection:graphstruct}

If the reader is familiar with graph structures in general, feel free to move on to the next sections. It is always possible to jump back here if necessary.

\vspace{2mm}

Chordal graphs, called also triangulated or decomposable, are of
great interest in many areas of mathematics. 

In this paper we relate them to a special kind of junction trees, therefore
a short introduction is needed.

Consider a graph $G=G(V,E)$ with the set of vertices $V = \{v_1,\dots,v_n\}$ and the set of undirected
edges $E$.

\begin{definition}
A subset of vertices $U\subseteq V$ defines an \textit{induced subgraph of }$%
\mathit{G}$ which contains all the vertices $U$ and any edges in $E$ that
connect vertices in $U$. 
\end{definition}

\begin{definition}
A subgraph induced by $U\subseteq V$ is called clique if it is \textit{%
complete} i.e. all pairs of vertices in $U$ are connected in $G$. 
\end{definition}

\begin{figure}
	\centering
	\includegraphics[width=1.0\textwidth]{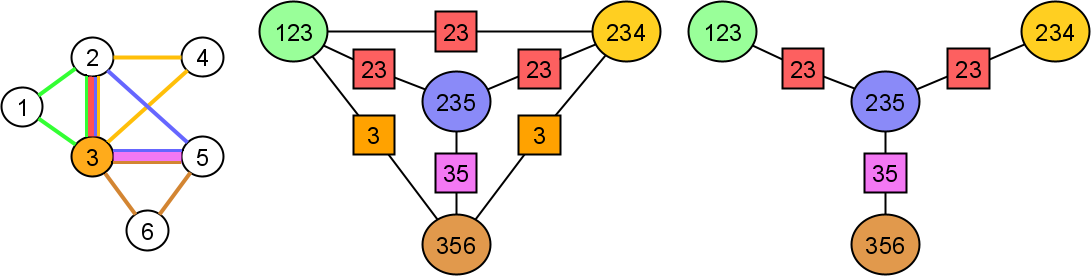}
	\vspace{-5mm}
	\caption{Example for the construction of chordal to junction (cherry) tree. On the left, the maximal  clusters (here, of size $3$) are denoted by different colors. In the middle, we changed to the cluster notation. The intersections (separators) are shown in squares. If each edge is weighted by the cardinality of its separator, then a maximal weight spanning tree becomes a junction (cherry) tree. One potential outcome is on the right.}
	\label{chordal_to_cherry_example}
\end{figure}

\begin{definition}
A graph $G\left( V,E\right) $ is said to be chordal when every cycle of
length $4$ or more has a chord (an edge joining two non-consecutive vertices
of the cycle). (See the leftmost graph on Figure \ref{chordal_to_cherry_example}.)
\end{definition}

\begin{definition}
Given a graph $G = (V,E)$ and a node $v \in V$, the neighbourhood of $v$ is defined as

$$\text{Ne}(v) = \{w \in V | (v,w) \in E\}$$

or all the nodes in $G$ that connect to $v$.
\end{definition}

\begin{definition}
The perfect elimination ordering of a graph $G(V,E)$ is an ordering $r_1,\dots,r_n$ of its vertices $v_1,\dots,v_n$ such that for all $i \in \{1,\dots,n\}: \ \text{Ne}(r_i) \cap \{r_{i+1},\dots,r_n\}$ is a clique in the remaining subgraph of $G(r_{i+1},\dots,r_n)$.
\end{definition}

Not all graphs $G$ have a perfect elimination ordering. The following theorem gives a necessary and sufficient condition for this property.

\begin{theorem}
$G$ is chordal if and only if $G$ has a perfect elimination ordering. \cite{rose1970triangulated}
\end{theorem}

Basic concepts and properties of chordal graphs can be found in \cite{golumbic2004algorithmic} and \cite{blair1993introduction}.

We need the following basic definition. 

\begin{definition}
A maximal clique of a graph $G$ is a clique which is not a subgraph of any
other clique of $G$. We will call these clusters.
\end{definition}

We will now introduce the general concept of an intersection graph. \cite{gavril1974intersection}

\begin{definition}
Consider a family of non-empty sets. The intersection graph of this family is obtained by representing each set by a vertex, two vertices are connected by an edge if and only if the coresponding sets intersect. (See the central graph on Figure \ref{chordal_to_cherry_example}.)
\end{definition}

The problem of characterizing the intersection graph of a family of sets having a defined topologic pattern is of great interest in different domains (for example interval graphs).

Let us suppose we have a graph $G$. We denote the set of clusters
(maximal cliques) by $\mu (G)$ and the set of clusters which contain the
vertex $v\in V$ by $\mu _{V}(G)$.

\begin{theorem}\label{theorem:subtree_graph}
A graph $G(V)$ is a subtree graph if and only if there exists a tree $T$
whose set of vertices is $\mu (G)$, so that, for every $v\in V$,$T(\mu
_{V}(G))$ is connected. \cite{gavril1974intersection}
\end{theorem}

\begin{theorem}\label{theorem:subtree_is_chordal}
$G$ is a subtree graph if and only if it is chordal graph. \cite{gavril1974intersection}
\end{theorem}

The next result was inspired by a definition given in \cite{thomas2009enumerating}.

\begin{definition}
Weigthed cluster- intersection graph of a chordal graph is an intersection
graph with the set of vertices defined by $\mu (G)$, whose edges connect
the non-disjoint clusters. The weight of the edges is given by the
cardinality of the elements of the clusters it connects.
\end{definition}

\begin{theorem}
A maximum weighted spanning tree of the weighted cluster intersection graph
is a characterisation of the chordal graph $G$.
\end{theorem}

\begin{definition}
Every maximum-weight spanning tree of the cluster-intersection graph of $G$
is called a cluster-tree of $G$. (See the rightmost graph on Figure \ref{chordal_to_cherry_example}.)
\end{definition}

We can define the junction tree in the following way.

\begin{definition}
\label{def:junction_tree}
A junction tree is obtained when the clusters are the maximal cliques of a chordal graph, and the edges, called separators, are given by the set of elements in the intersection of the endpoint clusters of the given edge.
\end{definition}

For any junction tree the following property, called running intersection
property holds.

\begin{theorem}
If an element is contained in two different clusters of a junction tree, then is contained in all separators and clusters on the path between the two clusters.
\end{theorem}

This theorem follows straightforward the Theorem \ref{theorem:subtree_is_chordal}.

Now we can define the $k$-th order cherry tree introduced in (Sz\'{a}ntai-Bukszar, Kovacs-Sz\'{a}ntai) as a special junction tree with clusters of size $k$ ($k$ elements in the cluster) and separators of size $k-1$. \cite{cherry1}

\begin{definition}
A $k$-order cherry tree is a special junction tree, in which all clusters consist of $k$ elements, and all separators consist of $k-1$ elements. (See the rightmost graph on Figure \ref{chordal_to_cherry_example}.)
\end{definition}

\subsection{Copula functions}\label{section:copula_functions}

In many applications, it is common to work with high-dimensional datasets. Our goal is often to approximate the joint cumulative distribution function (c.d.f.) and the joint probability distribution function (p.d.f.) of the underlying probability distribution of this data.

To solve this problem, Abe Sklar has introduced copula functions in 1959 \cite{sklar}, which are capable of modeling the connections between variables and their one-dimensional marginals.

Let $\boldsymbol{X} = (X_1,\dots,X_n) \in \mathbb{R}^n$ be an $n$-dimensional (continuous) random vector, with $f$ joint p.d.f. and $F$ joint c.d.f., meaning that

$$F(x_1,\dots,x_n) = \mathbb{P}(X_1 \le x_1, \dots, X_n \le x_n)$$

and

$$f(x_1,\dots,x_n) = \frac{\partial^n F(x_1,\dots,x_n)}{\partial x_1,\dots,\partial x_n}$$

Note: If $\boldsymbol{X}$ is a discrete random variable, then instead of an $f$ p.d.f. we can use its simple probability distribution: $\mathbb{P}(X_1 = x_1, \dots, X_n = x_n)$.

$\forall i \in \{1,\dots,n\}$, let the one-dimensional marginals of $F$ and $f$ be:

$$F_i(x_i) = \mathbb{P}(X_i \le x_i) = \int_{-\infty}^{x_i} f_i(t) dt$$

$\mathbb{P}(X_i \le x_i)$ can be easily approximated from the data, using the empirical c.d.f.

Let us denote the dataset by $\mathfrak{X} \in \mathbb{R}^{m \times n}$, $\mathfrak{X} := [x_{i,j}]_{i \in \{1,\dots,m\},j \in \{1,\dots,n\}}$, where each row can be thought of as a sample from the underlying distribution of $\boldsymbol{X} \in \mathbb{R}^n$.

Now we will introduce the definition of the copula function.
Since $F_i$ is the c.d.f. of $X_i$, the distribution of $F_i(X_i)$ is uniform on $[0,1]$. This can easily be seen from:

$$\mathbb{P}(F_i(X_i) \le x_i) = \mathbb{P}(X_i \le F^{-1}(x_i)) = F(F^{-1}(x_i)) = x_i$$

\begin{definition} A function $C: [0,1]^n \to [0,1]$ is an $n$-dimensional copula function, if it satisfies the following properties:

\begin{itemize}
\item $C(u_1,\dots,u_n)$ is strictly increasing in all $u_i$ components.
\item $C(u_1,\dots,u_{i-1},0,u_{i+1},\dots,u_n) = 0$ for all $u_k \in [0,1]$, $k \ne i$, $i \in \{1,\dots,n\}$.
\item $C(1,\dots,1,u_i,1,\dots,1) = u_i$ for all $u_i \in [0,1]$, $i \in \{1,\dots,n\}$.
\item $C$ is $n$-decreasing, meaning that for all $(u_{1,1},\dots,u_{1,n})$ and $(u_{2,1},\dots,u_{2,n})$ in $[0,1]^n$, if for all $i$, $u_{1,i} < u_{2,i}$, then

$$\sum_{i_1 = 1}^2 \cdots \sum_{i_n = 1}^2 (-1)^{\sum_{j=1}^n i_j} C(u_{i_1,1},\dots,u_{i_n,n}) \ge 0$$
\end{itemize}

\end{definition}

Let $(U_1,\dots,U_n) = (F_1(X_1),\dots,F_n(X_n))$, where all the $U_1,\dots,U_n$'s have $\text{UNI}[0,1]$ distribution separately. Then the $C$ copula function can be defined the following way:

$$C(u_1,\dots,u_n) = \mathbb{P}(U_1 \le u_1,\dots,U_n \le u_n)$$

which is the joint c.d.f. of $(U_1,\dots,U_n)$.

In the case where $X_1,\dots,X_n$ were independent, the joint distribution of the $U_i$'s will simply be a multivariate uniform distribution. However in the case where $X_1,\dots,X_n$ were not independent, we could get a different distribution. Because of this, $C(u_1,\dots,u_n)$ "expresses the correlations" between $(X_1,\dots,X_n)$.

In the theory of copulas, Sklar's theorem (1959) can be regarded as a central theorem \cite{sklar}:

\begin{theorem}
\label{sklar_theorem} Any multivariate c.d.f. $F$ can be written up the following way:

$$F(x_1,\dots,x_n) = C(F_1(x_1),\dots,F_n(x_n))$$

Moreover, if the one-dimensional marginals $F_1,\dots,F_n$ are continuous, then $C$ in unique.
\end{theorem}

Using Sklar's theorem, the $C$ copula c.d.f. can be calculated in the following way:

$$C(u_1,\dots,u_n) = F(F^{-1}_1(u_1),\dots,F^{-1}_n(u_n))$$

And using the chain rule, it is also true that

\begin{equation}\label{eq:joint_f}
f(x_1,\dots,x_n) = c(F_1(x_1),\dots,F_n(x_n)) \cdot f_1(x_1) \cdot \dots \cdot f_n(x_n)
\end{equation}

Where $c$ is the joint p.d.f. of $\boldsymbol{U} = (U_1,\dots,U_n)$, or the derivative of $C$. Meaning that a multivariate p.d.f. can be written up as the product of the one-dimensional marginal p.d.f.'s and a copula function. We will use this in the following section.

\subsection{Vine copulas}\label{section:vine_copulas}

A copula function can typically be described using 1, 2 or 3 parameters, which is not always capable of describing the connections between variables in higher dimensions. In 2001, Bedford T. and R. M. Cooke showed that $c(F_1(x_1),\dots,F_n(x_n))$ can be split into a special product, whose elements are pair-copulas and conditional pair-copula p.d.f.'s \cite{bedford1}. This formula was assigned to a specific graph structure, made up of a sequence of trees.

Before we present this formula, we need to show this specific graph structure, in which pair- and conditional pair-copula p.d.f.'s appear in a structured manner. Bedford T. and R.M. Cooke called this structure a "vine", which was explained in detail in their 2002 paper \cite{bedford2}. In their Definition of Regular vine on Page 1042, they used trees that can have nodes with cardinality $\ge 1$. From now on we will refer to these nodes as "clusters". Note that \cite{aas_dvine} also keeps the "node" usage. We will refer to trees that have nodes with cardinality $\ge 1$ as "cluster trees".

The vine structure is a special sequence of cluster trees. The first cluster tree is an ordinary tree, whose vertices are the indices of the variables ($1,\dots,n$). The graph structure on $n$ variables contains a total of $n-1$ cluster trees.

Let us denote the $k$'th cluster tree in the sequence by $T_k$. These cluster trees are defined using the following rules:

\begin{itemize}
\item  $T_1$ is any spanning tree on vertices $1, \dots, n$.
\item The $T_k$ cluster tree is defined by $n-k+1$ sets (clusters). Each cluster in $T_k$ contains exactly $k$ elements: $\{a_1,\dots,a_{k}\}$. To keep it simple, when drawing the tree, we will omit the set notation.
\item If $A = a_1\dots a_k$ and $B = b_1\dots b_k$ are two connected clusters, then the label of the edge running between them should be $D|S$, where $D = (A \cup B) \setminus (A \cap B)$, also known as the symmetric difference of sets $A$ and $B$, and $S = A \cap B$, or the intersection of sets $A$ and $B$. (For example if $A = 235$ and $B = 236$, then $D|S = 56|23$. If $S = \emptyset$, then we omit $S$, and also omit the line in front of it.)
\item The clusters of $T_{k+1}, k \ge 1$ contain exactly the same elements as the edge labels of $T_k$. (For example if the edge labels of $T_3$ are $14|23$ and $25|34$, then the clusters of $T_4$ are $1423$ and $2534$.)
\item Two clusters ($A$ and $B$) can only be connected if $|(A \cup B) \setminus (A \cap B)| = 2$. Because of this, in the label of the connecting edge $D|S$, $D$ will contain exactly $2$ elements.
\end{itemize}

Any $T_k$ tree defined by the previous rules is a cluster tree, however for simplicity, and if it does not cause confusion, we will sometimes refer to them as trees.

Now that the vine structure has been defined, we will show how to assign pair-copulas and conditional pair-copulas to it. These will be used to define the joint copula p.d.f.

We can assign a (conditional) pair-copula to each edge of every tree in the vine structure: If $E_1 = e_1a_1\dots a_k$ and $E_2 = e_2a_1\dots a_k$ are two connected clusters, then the edge running between them has to have a label of $e_1e_2|a_1\dots a_k$ (using the properties above). Every $e_1e_2|a_1\dots a_k$ edge label gets assigned a $c_{e_1e_2|S}$ conditional pair-copula density function, where $S=a_1\dots a_k$. If $S = \emptyset$, then we simply assign (regular) pair-copula densities to these edges.

\begin{wrapfigure}{r}{0.25\textwidth}
	\centering
	\includegraphics[width=0.25\textwidth]{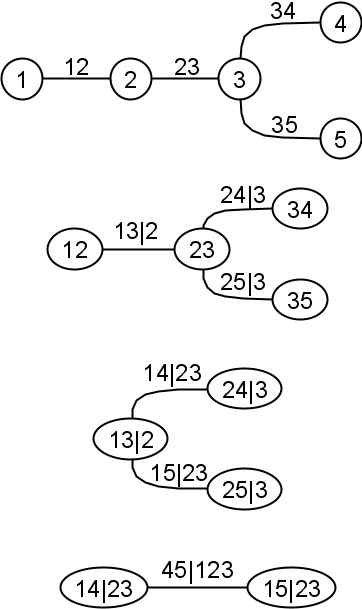}
	\caption{Example of a vine structure on 5 \mbox{variables}}
	\label{vine}
\end{wrapfigure}

For example if $A = 235$ and $B = 236$, then $D|S = 56|23$. The conditional pair-copula p.d.f. assigned to this edge is  $c_{56|23}$, which corresponds to the $56|23$ label on the edge.

Note: Clusters will not have a role in the output in this setup, so if it is more convenient, they can also be labeled in an $D|S$ format. This way, the previous edge label can simply be copied into the cluster, but $D \cup S$ is meant by it.

Vine structures are easy to draw (see Figure \ref{vine}). We start from a spanning tree, and we build every consecutive tree from the previous one, copy the edge labels into the new node labels, and connect up the resulting clusters so that they form a cluster tree. We also have to make sure to only connect clusters where the size of the symmetric difference of the labels  is exactly $2$. In the $k$'th tree $T_k$ this is equivalent to having an intersection of size $k-1$, since the symmetric difference and the intersection gives the whole union:

$$((A \cup B) \setminus (A \cap B)) \cup (A \cap B) = (A \cup B)$$

Therefore all clusters of the $k$'th tree will contain $k+1$ elements, and the label of each edge will contain $2$ elements before the condition line, and $k-1$ elements after.

So far we have defined the vine structure, as well as pair- and conditional pair-copulas assigned to the tree. Using these, according to Bedford T.'s and R.M. Cooke's 2001 theorem, any joint p.d.f. $f$ can be written up using conditional pair-copula density functions the following way.

Let us denote the set of edges in the $k$'th tree with $E_k$, and the conditional pair-copula p.d.f.'s that can be read from the edges with  $c_{e_1,e_2|S}$, where, in accordance with the above notation, $D = \{e_1,e_2\}$. Then any joint p.d.f. $f$ can be written up with pair-copula p.d.f.'s the following way \cite{bedford1}:

\begin{equation}
\label{joint_f2}
f_{\theta}(\vb{x}) = \prod_{j=1}^n \underbrace{f_j(x_j)}_{\begin{matrix}[0.5]\text{\scriptsize one-dimensional}\\ \text{\scriptsize marginals}\end{matrix}} \prod_{i=1}^{n-1} \prod_{(e_1,e_2|S) \in E_i} \underbrace{c_{e_1,e_2|S}}_{\begin{matrix}[0.5]\text{\scriptsize conditional pair-copulas}\\ \text{\scriptsize in the vine structure}\end{matrix}}\underbrace{(F_{e_1|S}(x_{e_1}|\vb{x}_S),F_{e_2|S}(x_{e_2}|\vb{x}_S),\vb{\theta})}_{\begin{matrix}[0.5]\text{\scriptsize conditional}\\ \text{\scriptsize c.d.f.'s}\end{matrix}}
\end{equation}

Where $\theta$ is the parameter of the pair copula density, and $F_{e|T}(x_e|\vb{x}_S)$ is a conditional c.d.f. This is usually not easy to approximate directly from data, however Joe (1996) showed that the conditional c.d.f.'s inside the $c$ function can be calculated the following way \cite{joe}:

\begin{equation}
\label{joe}
F_{e|S}(x_e|\vb{x}_S) = \frac{\partial C_{ej|S \setminus \{j\}}(F_{e|S \setminus \{j\}}(x_e|\vb{x}_{S \setminus \{j\}}),F_{j|S \setminus \{j\}}(x_{j}|\vb{x}_{S \setminus \{j\}}))}{\partial F_{j|S \setminus \{j\}}(x_{j}|\vb{x}_{S \setminus \{j\}})},
\end{equation}

\begin{wrapfigure}{r}{0.35\textwidth}
	\centering
	\includegraphics[width=0.35\textwidth]{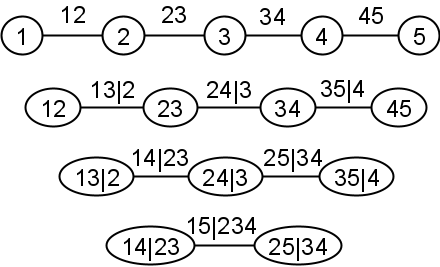}
	\caption{Example for a D-vine structure on 5 variables}
	\label{d_vine}
    \vspace{-1cm}
\end{wrapfigure}

where $j$ is an element of $S$.

This formula helps because $F_{e|S \setminus \{j\}}$ and $F_{j|S \setminus \{j\}}$  have already appeared in the previous tree of the vine structure (if the correct $j$ is selected), so the $F$ c.d.f. can be recursively calculated from the first couple of two-variable c.d.f.'s, which are easily approximated from the data.

In most applications so far, vine structures have been very simple. For example, D-vines, in which all cluster trees are lines (see Figure \ref{d_vine}), or C-vines, in which all cluster trees are stars. A general vine structure is called an R-vine (or \textit{regular} vine).

\section{Graphical representations of regular vines}

In this section, we will introduce two vine representations, and show the correspondance between them.

\subsection{Vine structure representation using a cherry tree sequence}\label{subsection:vinecherryrepr}


The later portion of this paper will focus on encoding the vine structure into a matrix, which uses various graphical representations. We will now show a new representation called a "cherry tree" sequence, introduced by Edith Kovács and Tamás Szántai \cite{kovacs_szantai2}, \cite{kovacs_szantai}.

We will use the following definition:

\begin{definition}\label{theorem:regular_cherry}
A $k$'th order cherry tree with the property that all its separators form a cherry tree of order $k-1$ is called \textit{regular cherry tree}.
\end{definition}


Now we will show how a cluster tree of a vine structure corresponds to a cherry tree:

\begin{itemize}
\item The clusters of the cherry tree correspond to the clusters of the vine cluster tree. We will not separate the part before and after the condition here.
\item The separating clusters, that appear on the edges of each cherry tree, are the intersection of two connected clusters, and are denoted in a rectangle. From the definition of vines, it is easy to see that if a cluster has $k$ elements, then the separating sets will have $k-1$ elements, since each connected pair of clusters differ in exactly one/one elements.
\item It is necessary to connect each pair of clusters with a separator (again, denoted with a rectangle) in order for the cherry tree and the cluster tree to uniquely correspond to each other. A separator can appear on multiple edges of the cherry tree.
\end{itemize}

\begin{figure}[h]
	\centering
	\includegraphics[width=0.5\textwidth]{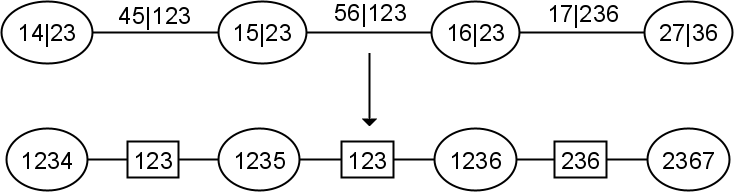}
	\caption{Creating a cherry three from a cluster tree of a vine structure}
	\label{junction_tree_repr}
\end{figure}

As for an example, refer to the tree on Figure \ref{junction_tree_repr}. Here, the sets $123$ and $236$ are the separators, but there are three clusters that join to $123$, since in the original graph these three clusters each have a connecting edge with a conditioning set of $123$.

The construction from a vine to a regular cherry tree sequence is unique forwards and backwards.

\begin{figure}[h]
	\centering
	\includegraphics[width=0.4\textwidth]{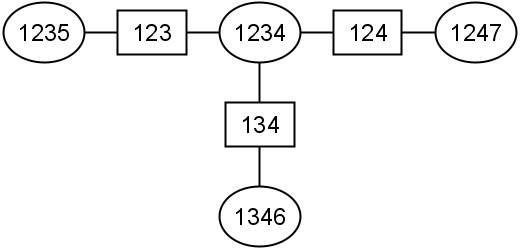}
	\caption{A 3rd order cherry tree which is not a regular cherry tree.}
	\label{irregular_cherry}
    \vspace{-2mm}
\end{figure}

Definition \ref{theorem:regular_cherry} is necessary, because vines cannot be represented by irregular cherry trees. So cherry trees are more general structures than trees inside vines. On Figure \ref{irregular_cherry}, one can see  a cherry tree which is not regular. (No matter how we try to form a cherry tree out of the clusters $123$, $124$ and $134$, the running intersection property will fail.)

The transformation of the original vine to the cherry tree sequence keeps the structure of each tree, and each pair of connected clusters in one of the original trees corresponds to a pair of connected clusters in the cherry tree, through a separator, which contains the intersection of the labels in the original clusters. The edge labels can therefore be constructed from the cherry tree, namely if clusters $A$ and $B$ are connected in the cherry tree with separator $S$, then the edge label in the original tree was $A \cup B \setminus S|S$.

\begin{theorem}\label{element_continuity} (Running intersection property)\\
For all cluster trees in the vine structure, if $A$ and $B$ different clusters in the same tree contain element $s$ (before or after the condition), then $s$ appears in all clusters and all edges on the path between $A$ and $B$.
\end{theorem}

\begin{pf}
We will now prove the statement for a specific cluster tree in the vine strucutre.

First of all, let us make it clear that in the $T_k$ tree of the vine structure, in every cluster, there are exactly the same number of elements, $k$. On the edge between two connected clusters there is a label of form $s_1 s_2 | t_1 \dots t_{k-1}$. So there are $k-1$ elements in the intersection of two connected clusters, and they differ in one/one elements (in order for us to obtain pair-copulas). In the cherry tree this fact is equivalent to all separators having size $k-1$.

Let $C_1, \dots, C_l$ be the clusters that are on the path between $A$ and $B$ in $T_k$. So $l \le k-2$, and all clusters are of size $k$. Let us indirectly assume that the statement is not true, so there exists an $i \in \{1,\dots,l\}$, such that $s \notin C_i$. We do not know anything about the clusters outside of $C_i$, so some may contain $s$, some may not. Let us now take the shortest path that contains $C_i$, but where $s$ does not appear in any cluster. Let us rename the endpoints of this path to $A$ and $B$. This way, we can split the problem into multiple smaller pieces, where $s \in A, s \in B$, but for all $i \in \{1,\dots,l\}$, $s \notin C_i$. It is sufficient to prove that we reach a contradiction for just one of these smaller chains (since if such chain cannot occur, then no chains of this type can occur, and it will not be possible to have a chain where some $C$ clusters contain $s$, but some do not):

\begin{figure}[h]
	\centering
	\includegraphics[width=0.7\textwidth]{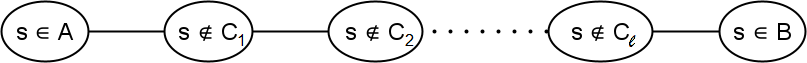}
	\label{proof1_lanc}
\end{figure}

From now on we will only work with this chain, and our goal will be to see, which part of $T_1$ it originated from. However first of all, let us count how many unique elements appear in the clusters of the above chain. $A$ contains $k$ different elements. $A$ and $C_1$ differ in one/one elements (for $A$ it's surely $s$), so we have obtained one new element with the $C_1$ cluster. $C_1$ and $C_2$ also differ in exactly one/one elements. (It is technically possible for the new element in $C_2$ to have appeared earlier, however, in that case, we could continue the proof with the 3-element chain of $A, C_1, C_2$ and get a contradiction that way. So we can assume that we always get new elements that we have not seen yet in the $C_i$ clusters that follow.) So we have obtained $1$ new element. And so on, up to $C_l$, we have obtained $l$ new elements. $C_l$ and $B$ differ in one/one elements once again, however for $B$ this is surely $s$, so we did not obtain a new element, since we have already counted $s$. So in this chain there are $k+l$ unique elements, and the chain consists of $l+2$ clusters.

Let us look at which elements could have appeared in the portion of $T_{k-1}$ where this chain originates from. There, the edges contain the exact same elements as $T_k$'s clusters. Since for all edge labels, the part before the condition and after the condition gives the union of elements in the clusters that the edge connects, there are also exactly $k+l$ elements in the portion of $T_{k-1}$ that the chain originates from. However, there are now $l+3$ elements on this chain.

And so on. We can work our way back all the way to $T_1$. Starting from $T_k$, we have made $k-1$ steps, so there are $l+2+k-1 = l+k+1$ nodes (one-element clusters) on this chain, but we can only place $k+l$ elements on the chain. So there is one more node in this chain of $T_1$, than the elements that we can place on this chain. However this is not possible, since all elements in the nodes of $T_1$ are unique.

Note: If we had allowed for $C_2$, $C_3$, $\dots$, or $C_l$ to contain no new elements, we would also have gotten a contradiction, since there would have been even fewer than $k+l$ elements to place in $l+k+1$ nodes.

With this, we have proven that there cannot be a chain in any tree of the vine structure that has $s$ in both of its endpoints, but not in between. So the statement of the theorem is true. $\square$
\end{pf}

\subsection{Vine representation using a chordal graph sequence}\label{subsubsection:chordalrepr}

Chordal graphs can also be used to represent a vine. This section will serve as the basis of Algorithm \ref{alg:vm2}, where the perfect elimination ordering property of chordal graphs will be used heavily. We will show how a chordal graph sequence can be constructed from a cherry tree sequence (which is a characterization of a vine structure).

\begin{definition}
A given cluster in a cluster tree is a leaf cluster, if it is connected to exactly one other cluster. The edge that runs between them is called a leaf edge.
\end{definition}

\begin{wrapfigure}{r}{0.25\textwidth}
	\centering
	\includegraphics[width=0.25\textwidth]{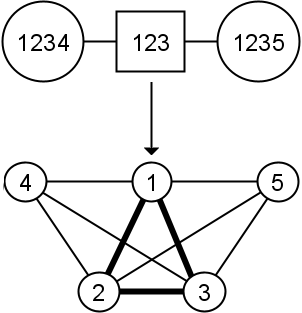}
	\caption{The chordal graph representation of two connected clusters. The separator, $123$ forms a complete graph (denoted by thicker lines), while the clusters $1234$ and $1235$ also form complete graphs.}
	\label{chordal_nodes_example}
    \vspace{-2cm}
\end{wrapfigure}

We will use the following transformation from a cherry tree to a new graph representation, which we later prove is chordal:

\vspace{1mm}

Every cluster $A = \{a_1,\dots,a_k\}$ is a complete graph on nodes $a_1,\dots,a_k$;

\vspace{1mm}

\hspace{-2mm}\begin{tabular}{ m{0.3cm} m{11cm} }
$\rightarrow$ & Every separator $S = \{s_1,\dots,s_{k-1}\} \subseteq \{a_1,\dots,a_k\} = A$ is then by the properties of the complete graph, also a a complete graph, but on $k-1$ nodes;
\end{tabular}

\vspace{1mm}

\hspace{-2mm}\begin{tabular}{ m{0.3cm} m{11cm} }
 $\rightarrow$ & This separator (intersection) of two clusters of size $k$ is a size $k-1$ clique. (See Figure \ref{chordal_nodes_example} for an example with $k=4$.)
\end{tabular}

\begin{theorem}\label{chordal_repr_is_chordal}
The graph obtained with the above method is chordal for any input cherry tree.
\end{theorem}

\begin{pf}
We will prove that the graph has a perfect elimination ordering. Let us start this ordering with one of the leaf clusters. Let us take the element which only appears in this leaf cluster. It follows from Theorem \ref{element_continuity} that such an element always exists. \textit{(Reductio ad absurdum)}

Repeat these steps until only one cluster remains in the cherry tree:

\begin{enumerate}
\item Find a leaf cluster in the cherry tree, and its corresponding clique in the graph.
\item Find the element which appears only in this clique.
\item Remove this leaf cluster from the cherry tree and this element from the graph.
\item Add this element to the perfect elimination ordering. This works, because this element connects to a clique (because of the construction of the graph).
\end{enumerate}

Then only one cluster remains. Write down all of the elements in this final cluster. Since it is part of a clique, all neighbourhoods of all of its nodes form cliques. This yields the perfect elimination ordering. $\square$
\end{pf}

\begin{theorem}\label{vine_perfect_elimination_ordering}
Given a vine on $n$ variables with cluster trees $T_1,\dots,T_{n-1}$ in chordal graph representation, there exists an ordering of its variables $1,\dots,n$, such that it is a perfect elimination ordering for every tree in $T_1,\dots,T_{n-1}$.
\end{theorem}

\begin{pf}
We will give a constructive proof to find this perfect elimination ordering.

Observation: Once an element became apart of a separator, it will remain in a separator in all subsequent cluster trees. This follows from the definition of vine structures.

Starting from $T_{n-1}$, the first two elements of the perfect elimination ordering should be the symmetric difference of the two clusters. Their neighbourhood is the separator of the two clusters, which forms a clique. According to the Observation, these elements could not have been in a separator in any previous cluster tree, so their neighborhood was always a separator of two clusters, which again forms a clique.

Moving on to $T_{n-2}$. If there are more elements that are not part of a separator, add them to the perfect elimination ordering. Once again, according to the Observation, they could not have come from a separator, so their neighborhood in any previous tree is an intersection of two clusters, which forms a clique.

And so on, moving backwards, add all elements in any order that are not part of a separator. According to the previous argument, this will always fulfil the conditions of a perfect elimination ordering.

After finishing, add all the remaining elements in any order. These have always formed a clique in every cluster tree, so any ordering of their elements is a valid perfect elimination ordering, thus the graph is chordal. $\square$
\end{pf}

We will conclude this section by introducing a new and important Definition that will be necessary for Section \ref{section:matrix}.

\begin{definition}\label{vine_peo}
The perfect elimination ordering of a vine structure is a perfect elimination ordering of all of its cluster trees. According to Theorem \ref{vine_perfect_elimination_ordering}., such an ordering always exists.
\end{definition}

In this section we have shown that the vine structure can be represented as a sequence of cherry trees and a sequence of chordal graphs. We have also proven multiple preliminary theorems, which will be expanded upon in the following parts.

\section{Matrix encoding of vine representations}\label{section:matrix}

In order to count the number of vine structures, vine matrices have been introduced \cite{napoles}, which were later used to store vines in a digital environment by \cite{dissmann}. These are, in all cases, $n \times n$ lower-triangular matrices, and their elements are the numbers $1, \dots, n$ that appear in the vine structure.

In this section, we will introduce two different vine matrix building methods, and later we will prove that under certain conditions they build the same matrix. Both methods take the graphical vine structure as an input. The first method corresponds to the one presented in paper \cite{napoles}. The second method is new, introduced by us, which differs from the first one in principle.

\subsection{O. M. Napoles' algorithm}\label{subsection:napolesalg}

Firstly we will prove two theorems that will be used later for the matrix representation. The following theorem refers to the cluster trees of the original vine representation.

The following theorem deals with not only the cluster trees that make up a vine structure, but the entire vine structure.

\begin{theorem}\label{leaf_nodes}
If the label on the edge of the final, $n-1$'st tree in the vine structure is $s_1 s_2 | t_1 \dots t_{n-2}$, then $s_1$ and $s_2$ can only appear in leaf clusters in any of the other trees, and $s_1$ and $s_2$ always appear before the condition.
\end{theorem}

\begin{pf} We will prove the statement for $s_1$. Afterwards it will clearly also be true for $s_2$, since there is no strict order for the elements before (or after) the condition.

Let the vine structure consist of $n-1$ trees. Let us assume that the statement does not hold for $T_{n-2}$, so the following holds:

\begin{figure}[h]
	\centering
	\vspace{-2mm}
	\includegraphics[width=0.5\textwidth]{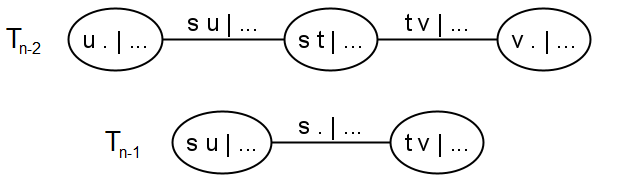}
	\vspace{-2mm}
	\label{proof2}
\end{figure}

Where $s$ appears on the edge of $T_{n-1}$, but in $T_{n-2}$ it does not appear in a leaf cluster.

Clearly $s \ne t$, since there are two different elements in every cluster of $T_{n-2}$ before the condition. $u$'s partner can only be $t$, and $v$'s partner can only be $s$ for $su$ and $tv$ to appear on the edges of $T_{n-2}$ before the condition. In this case in $T_{n-2}$, $t$ has to appear after the condition on the first edge, and $s$ has to appear after the condition on the second edge.

\begin{figure}[h]
	\centering
	\vspace{-2mm}
	\includegraphics[width=0.5\textwidth]{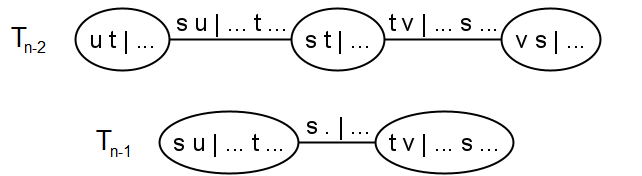}
	\vspace{-2mm}
	\label{proof3}
\end{figure}

But this is not possible, since then $s$ has to appear after the condition in $T_{n-1}$'s edge, because it appears in both of the clusters of $T_{n-1}$. Therefore we found a contradiction, so the original statement is true for $T_{n-1}$. Now we will prove the same for $T_k$, where $k \le n-2$. So in a general $T_k$ tree, if $s$ does not appear in a leaf cluster, then any of the following may happen:

\begin{itemize}
\item If $s$ appears in another cluster, then due to Theorem \ref{element_continuity}, it appears everywhere in the path between them. Therefore in the next step, $s$ will appear after in the intersection (after the condition), and in $T_{n-1}$, it will no longer be before the condition.
\item If $s$ does not appear elsewhere, only in a non-leaf cluster, then it is connected to at least two other clusters, with edge labels $s . | ...$. Again, because of this, there will be $2$ $s$'s in the next tree, so because of the previous point, $s$ will no longer appear in $T_{n-1}$. (This step is only valid from $T_{n-2}$ to $T_1$, since $s$ will only disappear through the following two trees. That's why we had to prove the statement separately for $T_{n-1}$.)
\end{itemize}

So, if the edge label of the last tree is $s_1 s_2 | ...$, then in all prior trees there is exactly one $s_1$ and one $s_2$, both in leaf clusters, before the condition; since if they appear after the condition, then they can never go back before the condition. $\square$
\end{pf}

\begin{rmk}
Theorem \ref{leaf_nodes} stays true for all layers of the vine structure, not just the last one. However the following Algorithm will only use this weaker statement.
\end{rmk}

Now, using the previous Theorems, we can set up the algorithm. Let us denote the vine matrix by $M$, and its elements by $m_{i,j}$, $i, j \in \{1,\dots,n\}$.

Oswaldo Morales Nápoles used a lower triangular array to store an R-vine \cite{napoles}. The idea is to store the constraint set of an R-vine in columns of an $n$-dimensional lower triangular array. We will now detail this algorithm, and amend it by the following: we will choose the minimum index whenever there is a choice to make, so that the result is unique:

\begin{algorithm}[H]
\caption{Column-wise minimal index vine matrix building method inspired by O. M. Nápoles' method}\label{alg:vm1}
\textbf{Input:} The trees in the vine structure with an adjacency list
\begin{algorithmic}
\State All edges in every tree start \textit{unmarked}.
\For{$j=1$ to $n-1$}
	\State $s_1 s_2|t_1 \dots t_{n-j-1} := $ The unmarked edge of $T_{n-j}$ \Comment{Theorem  \ref{unique_first_element}}
	\State Mark the edge $s_1 s_2|t_1 \dots t_{n-j-1}$
	\State $(m_{j,j}, m_{j+1,j}) :=(\min(s_1, s_2),\max(s_1,s_2))$
	\For{$i=j+2$ to $n$}
		\State $p_1 p_2|q_1 \dots q_{n-i} := $ The unmarked edge of $T_{n-i+1}$ in which $m_{j,j} \in \{p_1,p_2\}$ \Comment{Theorem \ref{unique_second_element}}
		\State Mark the edge $p_1 p_2|q_1 \dots q_{n-i}$
		\If{$m_{j,j} = p_1$}
			\State $m_{i,j} := p_2$
		\Else
			\State $m_{i,j} := p_1$
		\EndIf
	\EndFor
\EndFor
\State $m_{n,n} := n$
\State All other elements of $M$ are $0$.
\end{algorithmic}
\textbf{Output:} The $M$ matrix
\end{algorithm}

\begin{theorem}\label{unique_first_element}
$T_{n-j}$ has exactly one unmarked edge when Algorithm \ref{alg:vm1} reaches the tree.
\end{theorem}

\begin{pf}
We mark an edge in every tree in every step of the outer for-loop. Once we get to $T_{n-j}$, which contains $j$ edges, $j-1$ have already been marked. $\square$
\end{pf}

\begin{theorem}\label{unique_second_element}
$T_{n-i+1}$ has exactly one unmarked edge where $m_{j,j} \in \{p_1,p_2\}$.
\end{theorem}

\begin{pf}
This follows from Theorem \ref{leaf_nodes}. Since $m_{j,j}$ is either $s_1$ or $s_2$, and we have shown that in all the earlier trees there is exactly one cluster where $s_1$ or $s_2$ can be found.

Let this be $s_1$ now. According to Theorem \ref{leaf_nodes}, this will always be a leaf cluster, so the edge connecting it to the remaining tree is a leaf edge. So the unmarked portion of the vine makes up a new, smaller vine, where $s_1$ no longer occurs in any of the trees (since we have removed the cluster and the edge where it appeared), and we have removed exactly one leaf edge in every tree (which does not separate any of the trees, so what remains is an orderly vine structure on one fewer variables). $\square$
\end{pf}

\begin{ex}\label{example1}
We will now show how to build up the vine matrix with Algorithm \ref{alg:vm1} using the structure shown on Figure \ref{vine}. Firstly, all edges start unmarked, and we start from the last tree. Here, $s_1 s_2|t_1 \dots t_{n-j-1} = 45|123$. Out of $s_1$ and $s_2$, the smaller is $4$, so $(m_{1,1},m_{2,1}) = (4,5)$. Afterwards, we will mark (remove) the edge with label $s_1 s_2|t_1 \dots t_{n-j-1}$.
\end{ex}

\begin{center}
\begin{tabular}{ c c c c } 
$\begin{bmatrix}
4 & & & & \\
5 & & & & \\
 & & & & \\
 & & & & \\
 & & & & \\
\end{bmatrix}$&
$\begin{bmatrix}
4 & & & & \\
5 & & & & \\
1 & & & & \\
 & & & & \\
 & & & & \\
\end{bmatrix}$&
$\begin{bmatrix}
4 & & & & \\
5 & & & & \\
1 & & & & \\
2 & & & & \\
 & & & & \\
\end{bmatrix}$&
$\begin{bmatrix}
4 & & & & \\
5 & & & & \\
1 & & & & \\
2 & & & & \\
3 & & & & \\
\end{bmatrix}$\\
\\
\includegraphics[width=0.23\textwidth]{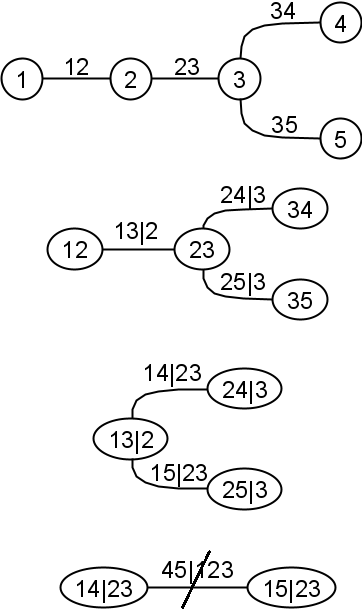}&
\includegraphics[width=0.23\textwidth]{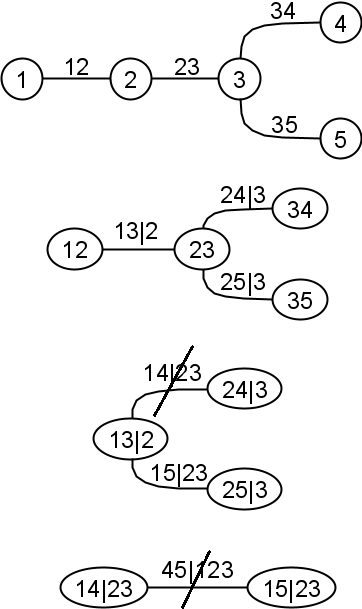}&
\includegraphics[width=0.23\textwidth]{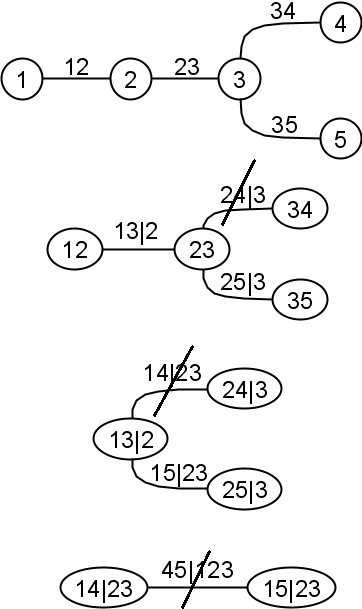}&
\includegraphics[width=0.23\textwidth]{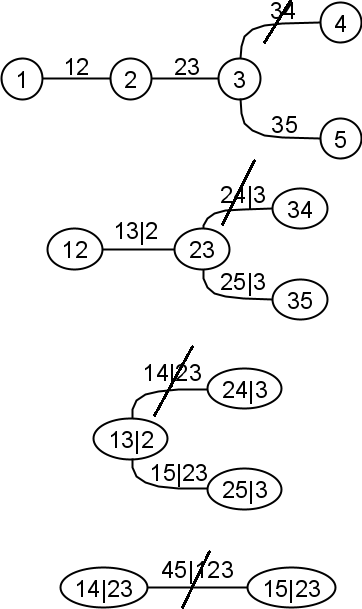}
\end{tabular}
\end{center}

Then we continue according to the previous figure through all trees in the vine structure, and in every step, we search for the number $4$ on the unmarked edge of each tree (or in the general case, for the number that we have written down on the top of the current column of the matrix). According to Theorem \ref{unique_second_element}, there is exactly one such unmarked edge in every tree, so every matrix element is unambiguous.

We delete the cluster tree that has all its edges marked. Afterwards, we continue by filling up the second column. Once again, from the bottom, we start from the last non-deleted tree, from its single unmarked edge ($15|23$). We choose the smaller number out of $1$ and $5$ (which is $1$), which we write down first, followed by the $5$:

\begin{center}
\begin{tabular}{ c c c c } 
$\begin{bmatrix}
4 & & & & \\
5 & 1 & & & \\
1 & 5 & & & \\
2 & & & & \\
3 & & & & \\
\end{bmatrix}$&
$\begin{bmatrix}
4 & & & & \\
5 & 1 & & & \\
1 & 5 & & & \\
2 & 3 & & & \\
3 & & & & \\
\end{bmatrix}$&
$\begin{bmatrix}
4 & & & & \\
5 & 1 & & & \\
1 & 5 & & & \\
2 & 3 & & & \\
3 & 2 & & & \\
\end{bmatrix}$\\
\\
\includegraphics[width=0.23\textwidth]{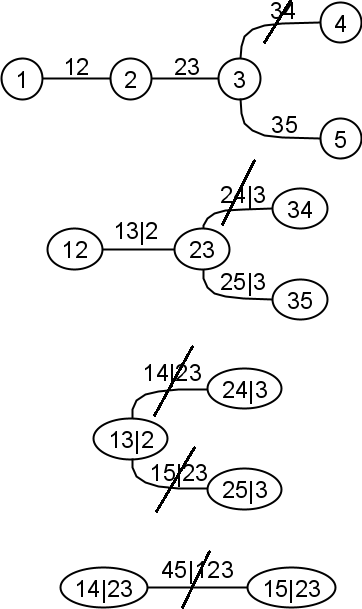}&
\includegraphics[width=0.23\textwidth]{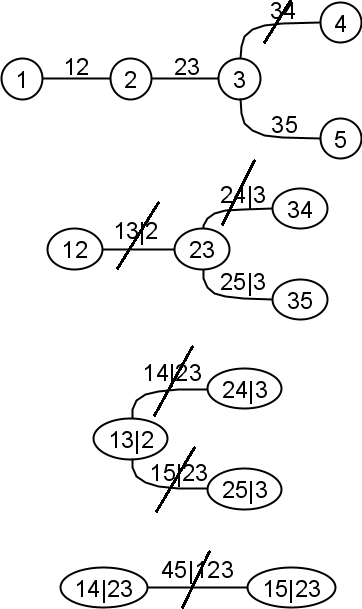}&
\includegraphics[width=0.23\textwidth]{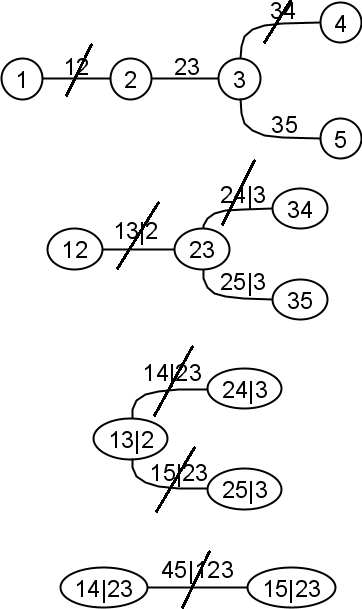}
\end{tabular}
\end{center}
Now we have deleted the second to last tree as well, since all (both) of its edges are marked. Let us once again start from the last non-deleted cluster tree, which has exactly one unmarked edge. First we insert the minimum on the unmarked edge, then its partner, then moving backwards in every other graph, we look for the given minimum value, and insert its partner into the matrix. The next figure shows the last two of such steps:

\begin{center}
\begin{tabular}{ c c c c } 
$\begin{bmatrix}
4 & & & & \\
5 & 1 & & & \\
1 & 5 & 2 & & \\
2 & 3 & 5 & & \\
3 & 2 & & & \\
\end{bmatrix}$&
$\begin{bmatrix}
4 & & & & \\
5 & 1 & & & \\
1 & 5 & 2 & & \\
2 & 3 & 5 & & \\
3 & 2 & 3 & & \\
\end{bmatrix}$&
$\begin{bmatrix}
4 & & & & \\
5 & 1 & & & \\
1 & 5 & 2 & & \\
2 & 3 & 5 & 3 & \\
3 & 2 & 3 & 5 & \\
\end{bmatrix}$\\
\\
\includegraphics[width=0.23\textwidth]{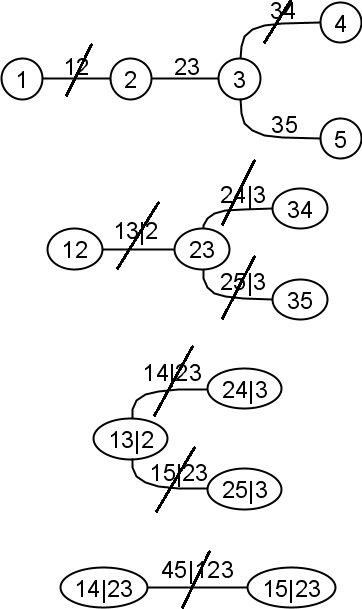}&
\includegraphics[width=0.23\textwidth]{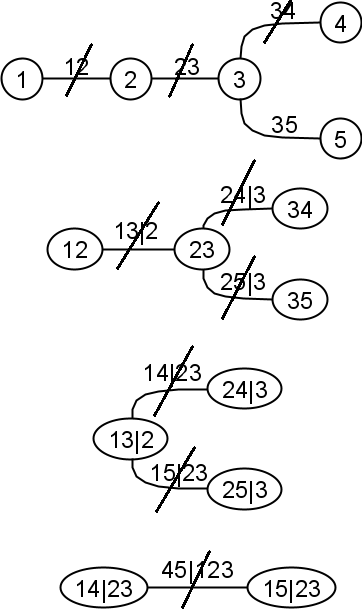}&
\includegraphics[width=0.23\textwidth]{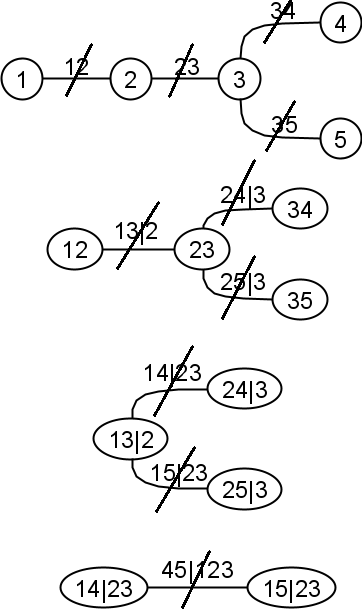}
\end{tabular}
\end{center}

And finally, we insert $n$ into $m_{n,n}$, or the bottom right element of the matrix, which in this case is $5$. This could never have appeared in the main diagonal, since we have always chosen the minimum out of two different elements, which could never have been the maximum, $5$. So the final matrix is:

$$M = \begin{bmatrix}
4 & & & & \\
5 & 1 & & & \\
1 & 5 & 2 & & \\
2 & 3 & 5 & 3 & \\
3 & 2 & 3 & 5 & 5 \\
\end{bmatrix}$$

In the O. M. Nápoles paper \cite{napoles} it is noted that the algorithm shown there could assign multiple matrices to a vine structure. So there is an option to define an equivalence class of matrices that contain all matrices which describe the same vine structure. This new, Algorithm \ref{alg:vm1} makes the matrix to vine structure assignment unique.

\begin{theorem}\label{alg1:runtime}
The number of comparisons in Algorithm \ref{alg:vm1} for an input vine on $n$ variables is
$$\frac{(n-1)(n(n+1)-3)}{3}$$
So the runtime of Algorithm \ref{alg:vm1} is $O(n^3)$.
\end{theorem}

\begin{pf}
The first step of the inner for loop has to find the first element (which was $4$ in our example) in every single tree, which takes $2\left(\sum_{k=1}^{n-1}k\right)-1 = n(n-1)-1$ comparisons, since this element can only appear before the constraint sign ($|$). The second inner for loop has to go through a vine which is one variable smaller, making it take $2\left(\sum_{k=1}^{n-2}k\right)-1 = (n-1)(n-2)-1$ comparisons.

And so on. Overall, until each edge is marked, we make the following number of comparisons: $$C:=(n(n-1)-1)+((n-1)(n-2)-1)+\dots+(3\cdot 2-1)+(2\cdot 1-1)$$

Then we can algebraically manipulate $C$ to obtain the formula given in the theorem:

\begin{align*}
C &= \sum_{r=1}^{n-1} \left((r+1)r-1\right) = \sum_{r=1}^{n-1} r^2 + \sum_{r=1}^{n-1} r - (n-1) = \frac{(n-1)n(2n-1)}{6}+\frac{(n-1)n}{2}-(n-1) = \\ &= (n-1)\left[\frac{n(2n-1)}{6}+\frac{3n}{6}-\frac{6}{6}\right] = \frac{n-1}{6}[2n^2-n+3n-6] = \frac{n-1}{3}[n(n+1)-3]
\end{align*}

Which is an $O(n^3)$ amount of comparisons. $\square$
\end{pf}

\subsection{Cherry tree algorithm}\label{subsection:cherryalg}

We present a new algorithm here for filling in the elements of an $M$ vine matrix. Additional theorems in this section will show that the algorithm indeed works.

\begin{algorithm}[H]
\caption{Row-wise vine matrix building method using a cherry tree sequence}\label{alg:vm2}
\textbf{Input:} The trees in the vine structure with an adjacency list
\begin{algorithmic}
\State Let $r_1,\dots,r_n$ be one of the vine's perfect elimination orderings defined in Definition \ref{vine_peo}. 
\Comment{Remark \ref{remark:alg1}, Theorem \ref{algo_main_diagonal}}
\For{$j=n$ to $1$}
	\State $m_{j,j} := r_j$
	\State $m_{n,j} := $ The node in $T_1$, where $r_j$ connects to one of $\{r_{j+1},\dots,r_{n}\}$ \Comment{Theorem \ref{one_cluster_T1}}
\EndFor
\For{$i=n-1$ to $2$}
	\For{$j=i-1$ to $1$}
		\For{$k=j+1$ to $i$}
			\State $A := \{m_{k,k}\} \cup \{m_{i+1,k},\dots,m_{n,k}\}$
			\State $B := \{m_{j,j}\} \cup \{m_{i+1,j},\dots,m_{n,j}\}$
			\If{the clusters $A$ and $B$ are connected in $T_{n-i+1}$} \Comment{Theorem \ref{node_appears}}
				\State $m_{i,j} = $ The single element of $A \setminus B$ \Comment{Theorem \ref{node_appears}}
			\EndIf
		\EndFor
	\EndFor
\EndFor
\State All other elements of $M$ are $0$.
\end{algorithmic}
\textbf{Output:} The $M$ matrix
\end{algorithm}

\begin{rmk}\label{remark:alg1}
Firstly let us examine why the perfect elimination order is needed here. We will once again work with the vine structure shown on Figure \ref{vine}, and assume that this is where we currently are in the algorithm:



$$
\begin{tikzpicture}[baseline=-\the\dimexpr\fontdimen22\textfont2\relax]
\matrix (m)[{matrix of math nodes},{nodes in empty cells},left delimiter={[},right delimiter={]},text=black]
{
4 & & & & \\
& 1 & & & \\
& & 2 & & \\
& \msquare & 5 & 3 & \\
3 & 2 & 3 & 5 & 5 \\
};
\begin{pgfonlayer}{myback}
\highlight[red]{m-2-2}{m-2-2}
\highlight[red]{m-5-2}{m-5-2}
\highlight[blue]{m-3-3}{m-3-3}
\highlight[blue]{m-5-3}{m-5-3}
\highlight[blue]{m-4-4}{m-5-4}
\end{pgfonlayer}
\end{tikzpicture}
$$
\nopagebreak
\begingroup
\setlength\arraycolsep{3pt}
$$\begin{matrix}
& \uparrow & \uparrow & \uparrow & \\
& B & A & A & \\
\end{matrix}$$
\endgroup

Where the upcoming element to fill is $\msquare$. For it to be unique, $A \setminus B$ has to contain exactly one element. Here $B = \{1,2\}$ and for $A$, we have the following two options: $\{2,3\}$ and $\{3,5\}$. Clearly, $A \setminus B$ has exactly one element if $A =\{2,3\}$, and then $A \setminus B = \{3\}$, so the new element is $\msquare = 3$.

The only way we can guarantee that the difference of these sets contains exactly one element is if the clusters $A$ and $B$ are connected in their cluster tree. We can only connect them if the symmetric difference of $A$ and $B$ contains $2$ elements, meaning they differ in one/one elements, $A \setminus B$ and $B \setminus A$. Looking at the vine structure it is easy to read that $12$ is not connected to $35$ in $T_2$, but it is connected to $23$:

\begin{figure}[h]
	\centering
	\includegraphics[width=0.6\textwidth]{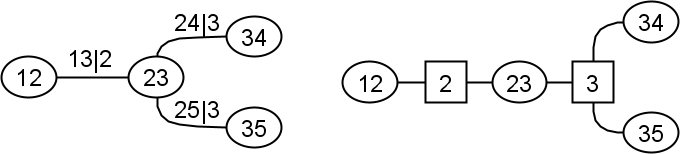}
	\label{vine_2}
\end{figure}

With the perfect elimination ordering of the vine; we get a reordeing of the indices $1,\dots,n$ so that this holds true. For every tree $T_k$, and every index in the perfect elimination ordering $r_j$, $\text{Ne}(r_j) \cap \{r_{j+1},\dots,r_{n-k+1}\}$ forms a complete graph, so $r_j$ is definitely connected to at least one cluster in $T_k$ containing at least one of $\{r_{j+1},\dots,r_{n-k+1}\}$. This way, there is at least one cluster $A$ for every cluster $B$ such that $A \setminus B$ contains exactly one element.

For now it may be unclear why such clusters even appear in the trees of the vine structure. This will be proven in Theorem \ref{node_appears}.
\end{rmk}

\begin{theorem}\label{algo_main_diagonal}
The main diagonal of the output of Algorithm \ref{alg:vm1} is also a perfect elimination ordering.
\end{theorem}

\begin{pf}
Let us examine $r_j$. In Algorithm \ref{alg:vm1}, this element is inserted into the matrix in the $j$'th step of the outer for-loop. Let us take tree $T_k$.

We need to show that $\text{Ne}(r_j) \cap \{r_{j+1},\dots,r_{n-k+1}\}$ forms a clique in every tree of the chordal representation of the vine sequence. Due to Theorems \ref{unique_second_element} and \ref{leaf_nodes}, when Algorithm \ref{alg:vm1} reaches $r_j$ in any tree, it will always be contained in exactly one leaf node, which connects to at most one other node. So when looking at the remaining elements $\{r_{j+1},\dots,r_{n-k+1}\}$, this $r_j$ neighbourhood is a single cluster, or a clique in the chordal representation. $\square$
\end{pf}

\begin{theorem}\label{one_cluster_T1}
In Algorithm \ref{alg:vm2}, in the construction of the main diagonal and the bottom row (in the first for-loop), $r_j$ connects to exactly one element out of $\{r_{j+1},\dots,r_{n}\}$ in $T_1$.
\end{theorem}

\begin{pf}
It connects to at least one node, since applying the perfect elimination ordering the same way as in \ref{remark:alg1} for $k = 1$, we get that $r_j$ connects to at least one node out of $\{r_{j+1},\dots,r_{n-k+1}\} = \{r_{j+1},\dots,r_n\}$.

But it cannot connect to any more, since if it did (say it connects to both $r_x$ and $r_y$), then there would exist an $r_j r_x ... r_y r_j$ cycle. (Since the remaining portion of the graph is connected, there exists a path between $r_x$ to $r_y$.)

And if $|\{r_{j+1},\dots,r_{n-k+1}\}| = 1$, for example $\{r_{j+1},\dots,r_{n-k+1}\} = \{r_x\}$, then $r_x$ has to once again connect to exactly one element in the remaining portion of the tree, since the tree is connected. $\square$
\end{pf}

\begin{theorem}\label{node_appears}
The clusters $A := \{m_{k,k}\} \cup \{m_{i+1,k},\dots,m_{n,k}\}$ and $B := \{m_{j,j}\} \cup \{m_{i+1,j},\dots,m_{n,j}\}$ appear in $T_{n-i+1}$, and if they are connected, then $A \setminus B$ contains exactly one element.
\end{theorem}

\begin{pf}
Let us start with $i = n-1$, just as Algorithm \ref{alg:vm2} does. Then, since all $m_{j,j} = r_{j}$ elements appear in $T_1$, and we chose $m_{n,j}$ to be the element that connects to $m_{j,j}$ in $T_1$, the edge label between the two nodes in $T_1$ is $m_{j,j} m_{n,j}$. These are exactly the elements that appear in the clusters of $T_2$, so the first half of the theorem is true for $i = n-1$.

Using induction, let us now decrease the value of $i$ in every step, and prove the theorem for $T_{n-i+1}$ using what we have proven for $T_{n-i}$. (For $i = n-1$, we have proven the theorem for $T_2$ using $T_1$, now we will prove it in order from $T_3, T_4, \dots$, etc. using the previous tree in each case.)

$T_{n-i}$ contains clusters of size $i$, so using the induction hypothesis we have shown that the clusters of form $A' = \{m_{k,k}\} \cup \{m_{i+2,k},\dots,m_{n,k}\}$, where $k \in \{j+1,\dots,i\}$ all appear in $T_{n-i}$. Let $A'$ be one of the previous sets for a fixed $k$, and $B' := \{m_{j,j}\} \cup \{m_{i+2,j},\dots,m_{n,j}\}$, which also appears in $T_{n-i}$ using the induction hypothesis. In order for the algorithm to deal with these sets, they have to be connected. But then, using the properties of the vine structure, they have to differ in exactly one/one element. So let us now rewrite them as $A' = \{s,p_1,\dots,p_{n-i-1}\}$ and $B' = \{t,p_1,\dots,p_{n-i-1}\}$. Then the label of the edge running between them is $s t | p_1 \dots p_{n-i-1}$, so in the following tree, $T_{n-i+1}$, the cluster $\{s,t,p_1,\dots,p_{n-i-1}\}$ will definitely appear. Let us call this cluster $C \in T_{n-i+1}$.

Now let us examine, what we would insert into the matrix $M$ into position $m_{i+1,j}$. According to Algorithm \ref{alg:vm2}, $A' \setminus B'$ would be inserted there. This element, according to the above expansion, is $s$.

With this, we now actually know what the original $B$ set was. Since

\vspace{-4mm}
\begin{align*}
B &= \{m_{j,j}\} \cup \{m_{i+1,j},\dots,m_{n,j}\} = \{m_{j,j}\} \cup \{s,m_{i+2,j},\dots,m_{n,j}\} = \\ &= \{s\} \cup (\{m_{j,j}\} \cup \{m_{i+2,j}\dots,m_{n,j}\}) = \{s\} \cup B' = \{s\} \cup \{t,p_1,\dots,p_{n-i-1}\} = C
\end{align*}

So the set $B$, as a cluster, indeed appears in the original tree, it will be the cluster of $T_{n-i+1}$ that was obtained from the edge between $A'$ and $B'$ in $T_{n-i}$.

Since set $A$ is a special $B$-type set, which was obtained by taken a different column into account (the $j$'th instead of the $k$'th), the theorem is also true for $A$, no matter what value $k$ takes.

Because of induction, the theorem will be true for all $j$ indices, so the sets $A$ and $B$, which were obtained from the matrix, indeed appear in the original vine structure as clusters.

The second statement of the theorem is that the set $A \setminus B$ contains exactly one element. However, this is almost trivial, since we can only get to this branch of the Algorithm is $A$ and $B$ are connected in $T_{n-i+1}$. As always, the label of the edge running between them is $s_1 s_2 | t_1 \dots t_{n-i-1}$, where $A$ and $B$ differ in exactly one/one element, $s_1$ and $s_2$. So $A \setminus B$ is either $\{s_1\}$ or $\{s_2\}$, definitely a one-element set. $\square$
\end{pf}

\begin{ex}\label{example2}
As we have done for Algorithm \ref{alg:vm1}, we will now show the construction of the matrix through an example. For the sake of simplicity, let us use the same example as before. Additionally, since in Theorem \ref{algo_main_diagonal} we have already shown that starting from the main diagonal that Algorithm \ref{alg:vm1} outputs gives a perfect elimination ordering that Algorithm \ref{alg:vm2} can use, let us use this main diagonal obtained previously, which was $4,1,2,3,5$.
\end{ex}

Algorithm \ref{alg:vm2} consists of two phases. In the first phase, the main diagonal and the bottom row will be filled out; starting from the last element in the above order, and building up $T_1$ by checking which nodes connect to the elements we have already connected:

\begin{center}
\begin{tabular}{ >{\centering\arraybackslash} m{9.2cm} >{\centering\arraybackslash} m{3cm} >{\centering\arraybackslash} m{0.24\textwidth} } 
{In the first step, the current graph only consists of node $\{5\}$, and the new node, $3$, does connect to it.} &
$\begin{bmatrix}
4 & & & & \\
 & 1 & & & \\
 & & 2 & & \\
 & & & 3 & \\
 & & & 5 & 5 \\
 \end{bmatrix}$&\includegraphics[width=0.23\textwidth]{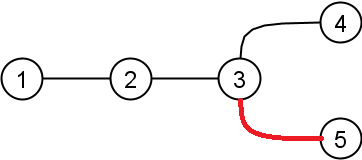}\\
{In the second step, the current graph now consists of nodes $\{3,5\}$, and the new node, $2$, connects to $3$.} &
$\begin{bmatrix}
4 & & & & \\
 & 1 & & & \\
 & & 2 & & \\
 & & & 3 & \\
 & & 3 & 5 & 5 \\
\end{bmatrix}$&\includegraphics[width=0.23\textwidth]{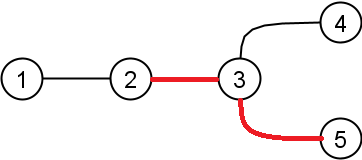}\\
\end{tabular}
\end{center}

\begin{center}
\begin{tabular}{ >{\centering\arraybackslash} m{9.2cm} >{\centering\arraybackslash} m{3cm} >{\centering\arraybackslash} m{0.24\textwidth} }
{In the third step, the current graph now consists of nodes $\{2,3,5\}$, and the new node, $1$, connects to $2$ out of these.} & 
$\begin{bmatrix}
4 & & & & \\
 & 1 & & & \\
 & & 2 & & \\
 & & & 3 & \\
 & 2 & 3 & 5 & 5 \\
\end{bmatrix}$&\includegraphics[width=0.23\textwidth]{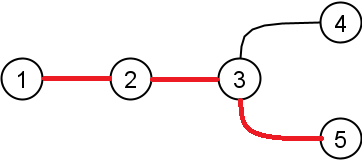}\\
{Finally, the current graph consists of nodes $\{1,2,3,5\}$, and the new node, $4$, connects to $3$ out of these.} & 
$\begin{bmatrix}
4 & & & & \\
 & 1 & & & \\
 & & 2 & & \\
 & & & 3 & \\
3 & 2 & 3 & 5 & 5 \\
\end{bmatrix}$&\includegraphics[width=0.23\textwidth]{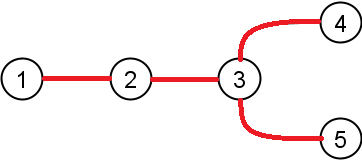}\\
\end{tabular}
\end{center}


Next up is the second phase of the algorithm, where all the remaining elements will be filled out. We will start from the bottom right, and move row by row. So the first element to fill out is $m_{4,3}$.  So, according to the algorithm, first $A = \{3\} \cup \{5\}$ (the known elements of column $4$), and $B$ will always equal to  $\{2\} \cup \{3\}$ in this step (the know elements of column $3$). In this step, $A$ will not even change, since $k$ goes from $j+1$ to $i$, but here $j = 3, i = 4$, so $k$ only makes one step.

\begin{wrapfigure}{r}{0.3\textwidth}
	\centering
	\vspace{-3mm}
	\includegraphics[width=0.3\textwidth]{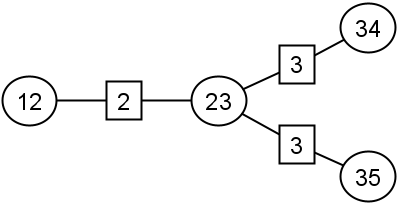}
\end{wrapfigure}

We will also have to check whether or not $A$ and $B$, as clusters, join in $T_{n-i+1} = T_2$. As we can see on the figure to the right, $35$ and $23$ do indeed join up, and this is fortunately guaranteed by the perfect elimination ordering set up at the start, as well as Theorems \ref{algo_main_diagonal} and \ref{node_appears}. So the new element is $\{3,5\} \setminus \{2,3\} = \{5\}$.

$$
\begin{tikzpicture}[baseline=-\the\dimexpr\fontdimen22\textfont2\relax ]
\matrix (m)[{matrix of math nodes},{nodes in empty cells},left delimiter={[},right delimiter={]},text=black]
{
4 & & & & \\
 & 1 & & & \\
 & & 2 & & \\
 & & 5 & 3 & \\
3 & 2 & 3 & 5 & 5 \\
};
\begin{pgfonlayer}{myback}
\highlight[black]{m-4-3}{m-4-3}
\highlight[red]{m-3-3}{m-3-3}
\highlight[red]{m-5-3}{m-5-3}
\highlight[blue]{m-4-4}{m-5-4}
\end{pgfonlayer}
\end{tikzpicture}
$$
\nopagebreak
\begingroup
\setlength\arraycolsep{3pt}
$$\begin{matrix}
& & \uparrow & \uparrow & \qquad\qquad\qquad\qquad\qquad\qquad\qquad\enspace \\
& & B & A & \qquad\qquad\qquad\qquad\qquad\qquad\qquad\enspace \\
\end{matrix}$$
\endgroup

Now let us move to the left in this row. $B$ is now $\{1\} \cup \{2\}$, and we have two options for $A$, either $A = \{2\} \cup \{3\}$ or $A = \{3\} \cup \{5\}$. In $T_2$, $12$ only connects to $23$, so $A \setminus B = \{2,3\} \setminus \{1,2\} = \{3\}$. So the new element is $m_{4,2} = 3$.

$$
\begin{tikzpicture}[baseline=-\the\dimexpr\fontdimen22\textfont2\relax ]
\matrix (m)[{matrix of math nodes},{nodes in empty cells},left delimiter={[},right delimiter={]},text=black]
{
4 & & & & \\
 & 1 & & & \\
 & & 2 & & \\
 & 3 & 5 & 3 & \\
3 & 2 & 3 & 5 & 5 \\
};
\begin{pgfonlayer}{myback}
\highlight[black]{m-4-2}{m-4-2}
\highlight[red]{m-2-2}{m-2-2}
\highlight[red]{m-5-2}{m-5-2}
\highlight[blue]{m-3-3}{m-3-3}
\highlight[blue]{m-5-3}{m-5-3}
\highlight[blue]{m-4-4}{m-5-4}
\end{pgfonlayer}
\end{tikzpicture}
$$
\nopagebreak
\begingroup
\setlength\arraycolsep{3pt}
$$\begin{matrix}
& \uparrow & \uparrow & \uparrow & \\
& B & A & A & \\
\end{matrix}$$
\endgroup

The last element in this row that we will fill out is $m_{4,1}$, which is calculated the following way: Here, $B$ is again always $B = \{4\} \cup \{3\}$, but now we have three options for $A$. Either $A = \{1\} \cup \{2\}$, or $A = \{2\} \cup \{3\}$, or $A = \{3\} \cup \{5\}$. In $T_2$, $34$ connects to $23$, so $A \setminus B = \{2,3\} \setminus \{4,3\} = \{2\}$, so the new element is $m_{4,1} = 2$.

$$
\begin{tikzpicture}[baseline=-\the\dimexpr\fontdimen22\textfont2\relax ]
\matrix (m)[{matrix of math nodes},{nodes in empty cells},left delimiter={[},right delimiter={]},text=black]
{
4 & & & & \\
 & 1 & & & \\
 & & 2 & & \\
2 & 3 & 5 & 3 & \\
3 & 2 & 3 & 5 & 5 \\
};
\begin{pgfonlayer}{myback}
\highlight[black]{m-4-1}{m-4-1}
\highlight[red]{m-1-1}{m-1-1}
\highlight[red]{m-5-1}{m-5-1}
\highlight[blue]{m-2-2}{m-2-2}
\highlight[blue]{m-5-2}{m-5-2}
\highlight[blue]{m-3-3}{m-3-3}
\highlight[blue]{m-5-3}{m-5-3}
\highlight[blue]{m-4-4}{m-5-4}
\end{pgfonlayer}
\end{tikzpicture}
$$
\nopagebreak
\begingroup
\setlength\arraycolsep{2pt}
$$\begin{matrix}
\uparrow & \uparrow & \uparrow & \uparrow & & & & \\
B & A & A & A & & & & \\
\end{matrix}$$
\endgroup

\begin{wrapfigure}{r}{0.17\textwidth}
	\centering
	\vspace{-12mm}
	\includegraphics[width=0.17\textwidth]{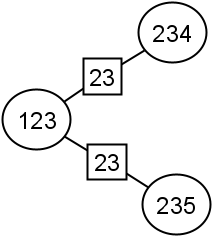}
\end{wrapfigure}

Next we will fill out the following row. The size of $A$ and $B$ increases by one, and we now have to look at $T_3$. First we will fill out $m_{3,2}$. $B$ is now $B = \{1\} \cup \{3,2\}$, and we only have one option for $A$, $A = \{2\} \cup \{3,5\}$. As you can see to the right, the nodes $123$ and $253$ are indeed connected in $T_3$, so the new element is $A \setminus B = \{2,3,5\} \setminus \{1,2,3\} = \{5\}$. Once again, Theorem \ref{node_appears} guarantees that this set always has one element.

$$
\begin{tikzpicture}[baseline=-\the\dimexpr\fontdimen22\textfont2\relax ]
\matrix (m)[{matrix of math nodes},{nodes in empty cells},left delimiter={[},right delimiter={]},text=black]
{
4 & & & & \\
 & 1 & & & \\
 & 5 & 2 & & \\
2 & 3 & 5 & 3 & \\
3 & 2 & 3 & 5 & 5 \\
};
\begin{pgfonlayer}{myback}
\highlight[black]{m-3-2}{m-3-2}
\highlight[red]{m-2-2}{m-2-2}
\highlight[red]{m-4-2}{m-5-2}
\highlight[blue]{m-3-3}{m-5-3}
\end{pgfonlayer}
\end{tikzpicture}
$$
\nopagebreak
\begingroup
\setlength\arraycolsep{3pt}
$$\begin{matrix}
& \uparrow & \uparrow & & & \\
& B & A & & & \\
\end{matrix}$$
\endgroup

And the last element in to fill out this row is $m_{3,1}$. Here $B = \{4\} \cup \{2,3\}$, and we have two options for $A$, either $A = \{1\} \cup \{2,3\}$, or $A = \{2\} \cup \{5,3\}$. In $T_3$, the cluster $243$ is connected to $132$, so $A \setminus B = \{1,2,3\} \setminus \{2,3,4\} = \{1\}$.

$$
\begin{tikzpicture}[baseline=-\the\dimexpr\fontdimen22\textfont2\relax ]
\matrix (m)[{matrix of math nodes},{nodes in empty cells},left delimiter={[},right delimiter={]},text=black]
{
4 & & & & \\
 & 1 & & & \\
1 & 5 & 2 & & \\
2 & 3 & 5 & 3 & \\
3 & 2 & 3 & 5 & 5 \\
};
\begin{pgfonlayer}{myback}
\highlight[black]{m-3-1}{m-3-1}
\highlight[red]{m-1-1}{m-1-1}
\highlight[red]{m-4-1}{m-5-1}
\highlight[blue]{m-2-2}{m-2-2}
\highlight[blue]{m-4-2}{m-5-2}
\highlight[blue]{m-3-3}{m-5-3}
\end{pgfonlayer}
\end{tikzpicture}
$$
\nopagebreak
\begingroup
\setlength\arraycolsep{3pt}
$$\begin{matrix}
\uparrow & \uparrow & \uparrow & & & & \\
B & A & A & & & & \\
\end{matrix}$$
\endgroup

\begin{wrapfigure}{r}{0.22\textwidth}
	\centering
	\includegraphics[width=0.22\textwidth]{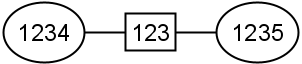}
\end{wrapfigure}

The final element which we have not filled out yet is $m_{2,1}$. Here $B = \{4\} \cup \{1,2,3\}$, and we have one option for $A$, $A = \{1\} \cup \{5,3,2\}$. The clusters corresponding to these sets are connected in $T_4$, as you can see on the right. So the new element is $A \setminus B = \{1,2,3,5\} \setminus \{1,2,3,4\} = \{5\}$. The fact that this has exactly one element is still guaranteed by Theorem \ref{node_appears}.

$$
\begin{tikzpicture}[baseline=-\the\dimexpr\fontdimen22\textfont2\relax ]
\matrix (m)[{matrix of math nodes},{nodes in empty cells},left delimiter={[},right delimiter={]},text=black]
{
4 & & & & \\
5 & 1 & & & \\
1 & 5 & 2 & & \\
2 & 3 & 5 & 3 & \\
3 & 2 & 3 & 5 & 5 \\
};
\begin{pgfonlayer}{myback}
\highlight[black]{m-2-1}{m-2-1}
\highlight[red]{m-1-1}{m-1-1}
\highlight[red]{m-3-1}{m-5-1}
\highlight[blue]{m-2-2}{m-5-2}
\end{pgfonlayer}
\end{tikzpicture}
$$
\nopagebreak
\begingroup
\setlength\arraycolsep{3pt}
$$\begin{matrix}
& \uparrow & \uparrow & & & & & & & \\
& B & A & & & & & & & \\
\end{matrix}$$
\endgroup

The matrix obtained is exactly the same as the one we have gotten after running Algorithm \ref{alg:vm1}. This is not an accident, as we will prove in the next section that if the output matrix has the same main diagonal in both cases, then the remaining elements will also be the same.

\begin{theorem}\label{alg2:graphwalk}
For all indices $i,j \in \{1,\dots,n\}$, each element of the matrix in Algorithm \ref{alg:vm2} can be obtained with a single traversal step on the $T_{n-i+1}$ cherry tree in the input vine sequence. The $i$'th row is obtained by fully traversing $T_{n-i+1}$ (the $i$'th tree counting backwards).
\end{theorem}

\begin{pf}
When filling our the bottom row, we are already traversing through $T_1$.

When filling out element $(i,j), i < n$ of the matrix with the $m_{i,j} := B \setminus A$ step, the algorithm considers elements in already filled out columns $j+1,\dots,n-1$ for the $A$ set (see the previous example). A set $A$ is chosen if $B$ and $A$ (as clusters) connect in tree $T_{n-i+1}$.

Now we will construct the corresponding graph traversal. Let us start traversing $T_{n-i+1}$ from the node defined by the initial $B$ set (the one that contains exactly $m_{j,j} \cup \{m_{i+1,j},\dots,m_{n,j}\}$), and step to a set $A$ chosen in the algorithm. When $j$ is decreased by $1$, let the new $A$ set be the node we are newly visiting and the new $B$ set the node we are visiting $A$ from. Repeat this for $j=i-1,\dots,1$, and we have traversed $T_{n-i+1}$, visiting each node with set $A$ exactly once. $\square$
\end{pf}

\begin{theorem}\label{alg2:runtime}
Given a perfect elimination ordering of the vine, if the elements inside each cluster and separator are ordered, then the number of comparisons in Algorithm \ref{alg:vm2} is
$$\frac{(n-1)n(n+4)}{6}$$
So with ordered elements the runtime of Algorithm \ref{alg:vm2} is $O(n^3)$. If the elements inside each cluster and separator are not ordered, then the runtime of Algorithm \ref{alg:vm2} is $O(n^3\log(n))$.
\end{theorem}


\begin{pf}
Firstly we walk through $T_1$ in $n$ steps, and write down the nodes we visited.

Then using the cherry tree traversal in the proof of Theorem \ref{alg2:graphwalk}, we can walk through the remaining trees as well. In $T_2$, there are $n-1$ nodes, so this takes $n-1$ steps. In $T_3$, there are $n-2$ nodes, so this takes $n-2$ steps, etc., in $T_k$ it takes $n-k+1$ steps to walk through the tree.

Then at each point of a walk, we have to calculate the difference of two equally sized sets of numbers. When the elements inside each cluster and separator are ordered, this can be done in linear time. For example if the cluster is stored as the list $A = [2,4,5,7]$, and the separator is stored as the list $B = [2,4,7]$, then the subtraction of $\{2,4,5,7\} \setminus \{2,4,7\}$ requires walking through both lists at the same time, and comparing them element-wise. If the compared elements are different, then the result will be the element inside the first list (the cluster's list).

If the elements inside clusters and separators are not ordered, then we can do the following: For each cluster or separator of size $n$, sort it in $O(n\log(n))$ steps. Since the algorithm goes through each cluster and separator exactly once, it will not sort any cluster or separator more than once.

Another option is to sum up the elements in the  cluster and subtract the sum of the elements in the separator:

$$\text{The single element of } A \setminus B = \sum_{a\in A}a - \sum_{b\in B}b$$

Since there are $n$ different elements in each cluster, their average size is no smaller than $O(\log(n))$. So adding up $n$ elements of size $O(\log(n))$ also takes $O(n\log(n))$ steps, which means this method is not faster than the sorting method.

So we have obtained that for $|A|=n,|B|=n-1$ calculating $A \setminus B$ takes $n$ steps when $A$ and $B$ are stored in ordered lists, and $O(n\log(n))$ steps when they are not.

In tree $T_k$, nodes are of size $k$, so subtracting elements of a separator from elements of a cluster (calculating $A \setminus B$) takes $k$ comparisons for each node in the ordered case, and $O(k\log(k))$ comparisons otherwise.

So in the ordered case we make $(n-k+1)k$ comparisons in $T_k$ overall, so the total number of comparisons that need to be made for Algorithm \ref{alg:vm2} is

$$C := \sum_{k=1}^{n-1} (n-k+1)k$$

We get obtain the formula in the theorem with simple algebraic manipulation:

\begin{align*}
C &= \sum_{k=1}^{n-1} (n-k+1)k = (n+1)\sum_{k=1}^{n-1} k - \sum_{k=1}^{n-1} k^2 = \frac{(n+1)n(n-1)}{2}-\frac{(n-1)n(2n-1)}{6} = \\ &= (n-1)n\left[\frac{3(n+1)}{6}-\frac{2n-1}{6}\right] = \frac{(n-1)n}{6}[3n+3-2n+1] = \frac{(n-1)n}{6}[n+4]
\end{align*}

Which is an $O(n^3)$ amount.

And in the unordered case an $O(n)$ operation is replaced by an $O(n\log(n))$ operation, so in that case, the algorithm makes $O(n^3\log(n))$ steps. $\square$
\end{pf}

\subsection{Equivalence of algorithms}\label{subsection:equivalence}

By definition, two algorithms are equivalent, if the same output is obtained every time the same input is given.

\begin{theorem}\label{equivalence}
Algorithms \ref{alg:vm1} and \ref{alg:vm2} are equivalent, if the main diagonals of the output matrices is the same. (The main diagonal is unique, if a perfect elimination ordering is preset.)
\end{theorem}

\begin{pf} 
Rephrasing the theorem, using the main diagonal of the output of Algorithm \ref{alg:vm1} at the beginning of Algorithm \ref{alg:vm2} makes it so Algorithm \ref{alg:vm2} generates the exact same matrix.

Because of this, $m_{j,j}, j \in \{1,\dots,n\}$ are the same in the output of both algorithms. Now we will show that $m_{n,j}, j \in \{1,\dots,n-1\}$, or the bottom row is also the same. (The last element of the bottom row, $m_{n,n}$, is part of the main diagonal, so it is definitely the same.)

$m_{n,j}$ is the element in Algorithm \ref{alg:vm2} with which $m_{j,j}$ connects to the subgraph $\{m_{j+1,j+1},\dots,m_{n,n}\}$ in $T_1$ (see Example \ref{example2}, first phase).

In Algorithm \ref{alg:vm1}, $m_{n,j}$ is generated in the $j$'th step of the outer for-loop such that we take the tree $T_{n-j} = T_{n-(n-1)} = T_1$, and check where $m_{j,j}$ appears on its unmarked edges before the condition, and we take its partner before the condition. In $T_1$, this is the entire edge label, namely $m_{j,j} m_{n,j}$. Since $T_1$ contains unique elements in its nodes, and they are one-element clusters, there can only be one edge where $m_{j,j} m_{n,j}$ occurs. This connects two nodes, $m_{j,j}$ and $m_{n,j}$, so $m_{j,j}$ indeed connects to the subgraph containing $m_{n,j}$. Now we only have to show that $m_{n,j} \in \{m_{j+1,j+1},\dots,m_{n,n}\}$. But this is obvious from the construction of Algorithm \ref{alg:vm1}. The elements of the $j$'th column can only be chosen out of the so far unused elements (so the ones not in $\{m_{1,1},\dots,m_{j,j}\}$), since according to Theorem \ref{unique_second_element} one of the numbers gets used up in every step, such that all edges and clusters that it appeared in get deleted.

With this we have proven that the bottom row is the same in the outputs of both algorithms. Now we will prove that the elements below the main diagonal, $m_{2,1},\dots,m_{n,n-1}$ are the same in both outputs.

In Algorithm \ref{alg:vm1}, these are exactly the partners of elements $m_{1,1},\dots,m_{n-1,n-1}$, which we get by the following: In every step $j$, look for $m_{j,j}$ before the condition on the unmarked edges of $T_{n-j}$, and take its partner. By construction, $m_{j,j} < m_{j+1,j}$, since $m_{j,j}$ was always the minimum of the two.

In Algorithm \ref{alg:vm2} we fill out these elements by already knowing $m_{j+2,j},\dots,m_{n,j}$ in the same column, and having $B = \{m_{j,j}\} \cup \{m_{j+2,j},\dots,m_{n,j}\}$. We also know that whichever cluster $B$ is connected to in $T_{n-j}$ (let us call this $A$) differs in exactly one/one element from $B$, and this new element will be $m_{j+1,j}$. Now let us examine the label of the edge running between $A$ and $B$. According to Theorems \ref{leaf_nodes} and \ref{unique_second_element}, one element gets used up in every step, $m_{j,j}$, which is deleted from the graph, and inserted into the main diagonal. This element cannot occur later, so it also cannot be part of $B$. So the label of the edge running between $A$ and $B$ has to be $m_{j,j},m_{j+1,j}|m_{j+2,j},\dots,m_{n,j}$, so Algorithm \ref{alg:vm2} would also insert the partner of $m_{j,j}$ which appears on an unmarked edge of $T_{n-j}$ into $m_{j+1,j}$. (And due to Theorem \ref{unique_second_element}, this element is unique.)

Now we will examine the $j$'th subdiagonal elements starting from $j = 2$ to $n-1$, namely elements $m_{1+j,1},\dots,m_{n,n-j}$.

In Algorithm \ref{alg:vm1}, these were generated similarly to $m_{2,1},\dots,m_{n,n-1}$, except instead of taking the partner of $m_{j,j}$ on the single unmarked edge of the last tree into consideration, the tree $i$ steps before that one is considered, namely $T_{n-j-i}$.

In Algorithm \ref{alg:vm2}, these were also generated similarly, except the sets $A$ and $B$ have $i$ fewer elements, and $T_{n-j-i}$ is considered here as well; specifically $B = \{m_{j,j}\} \cup \{m_{j+2+i,j},\dots,m_{n,j}\}$. There are $i+1$ options for $A$, $A = \{m_{j+k,j+k}\} \cup \{m_{j+2+i,j+k},\dots,m_{n,j+k}\}$, $k \in \{1,\dots,i\}$. Because of the perfect elimination ordering condition in Algorithm \ref{alg:vm2}, $B$ connects to at least one of these. Let us choose an $A$ that $B$ connects to by fixing a $k$ for $A$, so let $B$ and $\{m_{j+k,j+k}\} \cup \{m_{j+2+i,j+k},\dots,m_{n,j+k}\}$ be connected in $T_{n-j-i}$. We will again examine what the label of the edge running between $A$ and $B$ is. Once again due to Theorems \ref{leaf_nodes} and \ref{unique_second_element}, we use up an element in each step, so the elements in the main diagonal are unique, so $A \cap B$ is definitely $\{m_{j,j},m_{j+k,j+k}\}$. So the label of the edge running between $A$ and $B$ is $m_{j,j},m_{j+k,j+k}|m_{j+2+i,j},\dots,m_{n,j}$, and in this case, since $A$ and $B$ are connected, the remaining part has to be the same, namely $\{m_{j+2+i,j},\dots,m_{n,j}\} = \{m_{j+2+i,j+k},\dots,m_{n,j+k}\}$. So Algorithm \ref{alg:vm2} would insert $A \setminus B$ or $m_{j+k,j+k}$ into $m_{j+i+1,j}$.

This is indeed the same element that Algorithm \ref{alg:vm1} would insert into $m_{j+i+1,j}$, since according to the above, this is the partner of $m_{j,j}$ on the unmarked edges of $T_{n-j-i}$. (When Algorithm \ref{alg:vm1} deals with $T_{n-j-i}$, $\max(s_1,s_2)$ indeed becomes $m_{j+i+1,j}$.) The fact that $m_{j+i+1,j}$ is unmarked can be easily seen from this element appearing after the $j$'th column, so it was not yet used up until the $j$'th column, so it could not have appeared in the main diagonal until the $j$'th column, so it could not have been marked.

So both algorithms insert the same elements in each subdiagonal, as $j$ goes from $1$ to $n-2$, so each element is exactly the same in both output matrices. $\square$
\end{pf}

The equivalence can be observed through the previous Examples: \ref{example1} and \ref{example2}.

One may ask whether of not the matrix built up in Algorithm \ref{alg:vm2} is unique or not, or is there another perfect elimination ordering of the variables? The answer is that the matrix is not unique, and this follows from the construction of Algorithm \ref{alg:vm1}:

\begin{theorem}\label{number_of_matrices}
An $n$-variable vine has at least $2^{n-1}$ perfect elimination orderings.
\end{theorem}

\begin{pf}
We have proven that the main diagonal of the output of the first algorithm will always give a perfect elimination ordering (Theorem \ref{algo_main_diagonal}). The question is, how many ways there are to fill out the main diagonal.

In Algorithm \ref{alg:vm1} we have decided on the order $(m_{j,j},m_{j+1,j}) := (\min(s_1,s_2),\max(s_1,s_2))$ for all $j \in \{1,\dots,$ $n-1\}$. However any ordering or elements $s_1, s_2$ would have been valid, so any of the two elements before the condition could have been chosen on the last unmarked edge. These are $2$ possibilities in each step, and since we can choose independently in each step, there are overall $2^{n-1}$ possibilities. Since the variables do not have a strict order, the Algorithms could still run correctly if $s_1$ and $s_2$ were swapped. With this, a valid matrix can be generated by selecting any combination of the minimum or the maximum in each step. Any of these $2^{n-1}$ possibilities have to satisfy the perfect elimination ordering condition at the start of Algorithm \ref{alg:vm2}, due to Theorem \ref{algo_main_diagonal}. $\square$
\end{pf}

\section{Conclusions}\label{section:conclusions}

We gave rigorous graphical representations of vine copulas using various graph sequences. The original representation was introduced in \cite{bedford2}. Later an equivalent representation, given by a sequence of cherry trees was introduced in \cite{cherry1}. A third representation can be obtained using a chordal graph sequence. We have shown relationships between them.

Based on the original graphical structure, \cite{napoles} has introduced a way to represent vines in a lower-triangular matrix. Using this idea, we have constructed an algorithm that can fill out such a matrix column-wise, given a vine as its input. We have presented how it works through an example. Furthermore, we have introduced a new method to fill out the same matrix using the cherry tree sequence representation, and we have visualized it through the same example as well.

The two matrix filling methods are equivalent if the same vine perfect elimination ordering is given as an input. We have seen that such an ordering always exists.

We gave a theoretical background for the representations, equivalences and algorithms shown in this paper. The algorithms introduced will give the basis of further work in the field of vine copula constructions based on data.










\newpage
\printcredits

\bibliographystyle{cas-model2-names}

\bibliography{CIKK1}

\bio{}
\endbio


\end{document}